\pdfoutput=1

\documentclass[11pt]{article}
\usepackage[final]{acl}
\usepackage{times}
\usepackage{latexsym}
\usepackage[T1]{fontenc}
\usepackage[utf8]{inputenc}
\usepackage{microtype}
\usepackage{inconsolata}
\usepackage{graphicx}
\usepackage{xcolor}
\usepackage{enumitem}
\usepackage{booktabs}
\usepackage{array}
\usepackage{hyperref}
\usepackage{url}
\usepackage{multirow}
\usepackage{colortbl}
\usepackage{booktabs}
\usepackage{amssymb}
\usepackage{amsmath}
\usepackage{amsthm}
\usepackage{graphicx}
\usepackage{makecell}
\usepackage{threeparttable}
\usepackage{CJKutf8}
\usepackage{lscape}
\usepackage{fancyhdr}
\usepackage[ruled,vlined]{algorithm2e}

\usepackage{threeparttable}
\usepackage{wrapfig}
\usepackage{subfigure}
\usepackage{subcaption}
\usepackage{multirow}

\title{Unlocking Continual Learning Abilities in Language Models}

\author{
Wenyu Du$^1$\thanks{~~Equal Contributions.}\thanks{~~Work done during interning at NVIDIA.}
\quad Shuang Cheng$^{2*}$
\quad Tongxu Luo$^3$
\quad Zihan Qiu$^4$
\quad Zeyu Huang$^5$ \\
\bf{
Ka Chun Cheung$^6$
\quad Reynold Cheng$^1$\thanks{~~Corresponding Authors.}
\quad Jie Fu$^7$\footnotemark[3]}\\
$^1$School of Computing and Data Science, The University of Hong Kong \\
$^2$ICT, Chinese Academy of Sciences \quad $^3$CUHK-SZ \\
$^4$Tsinghua University \quad $^5$University of Edinburgh \quad $^6$NVIDIA \quad
$^7$HKUST \\
\texttt{wydu@cs.hku.hk \quad chengshuang22@mails.ucas.ac.cn}
}

\begin{document}
\maketitle
\begin{abstract}
Language models (LMs) exhibit impressive performance and generalization capabilities.
However, LMs struggle with the persistent challenge of catastrophic forgetting, which undermines their long-term sustainability in continual learning (CL). 
Existing approaches usually address the issue by incorporating old task data or task-wise inductive bias into LMs.
However, old data and accurate task information are often unavailable or costly to collect, hindering the availability of current CL approaches for LMs. 
To address this limitation, we introduce ``MIGU'' (\textbf{M}agn\textbf{I}tude-based \textbf{G}radient \textbf{U}pdating for continual learning), a rehearsal-free and task-label-free method that only updates the model parameters with large magnitudes of output in LMs' linear layers. 
MIGU is based on our observation that the L1-normalized magnitude distribution of the output in LMs' linear layers is different when the LM models deal with different task data. 
By imposing this simple constraint on the gradient update process, we can leverage the inherent behaviors of LMs, thereby unlocking their innate CL abilities. 
Our experiments demonstrate that MIGU is universally applicable to all three LM architectures (T5, RoBERTa, and Llama2), delivering state-of-the-art or on-par performance across continual finetuning and continual pre-training settings on four CL benchmarks. 
For example, MIGU brings a 15.2\% average accuracy improvement over conventional parameter-efficient finetuning baselines in a 15-task CL benchmark. 
MIGU can also seamlessly integrate with all three existing CL types to further enhance performance. 

\end{abstract}

\section{Introduction}

\begin{figure}
    \centering
    \includegraphics[width=\linewidth]{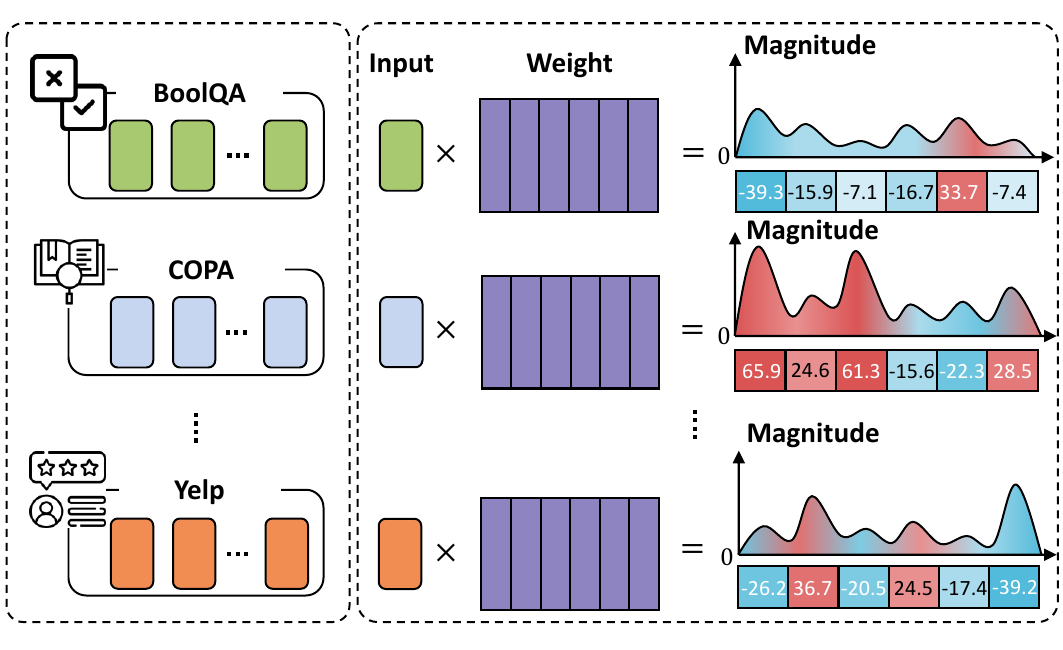}
    \caption{The different output values and the varying magnitude distributions are observed across the BoolQA, COPA, and Yelp datasets. The output values and L1-normalized magnitude distributions are from the real sample of the first linear layer in FFN of the last Transformer block of T5. The detailed magnitude distributions are illustrated in Figure~\ref{fig:smooth_vector_plot} of Appendix~\ref{visualization}.}
    \label{fig:motivation}
\end{figure}

Neural networks suffer from catastrophic forgetting~\cite{mccloskey1989catastrophic}, i.e. learning new knowledge and tasks at the cost of forgetting previously acquired ones. 
Recently, language models (LMs) have demonstrated impressive performance and generalization capabilities across the spectrum of NLP research~\cite{liu2019RoBERTa,brown2020language,touvron2023Llama} and beyond~\cite{damonlpsg2023videoLlama}. 
Nevertheless, they still suffer from catastrophic forgetting~\cite{shi2024continual,wu2024continual}, undermining the capacity for continual learning (CL)~\cite{wang2024comprehensive}. 
In light of the large scale and high cost of training LMs~\cite{achiam2023gpt}, models with strong continual learning capabilities would enable more economical reuse of these resource-intensive models, a vital trajectory for driving both scientific development and societal benefits.

To make LMs better continual learners, the research community pursues three main directions~\cite{shi2024continual}: 
(1) rehearsal-based approaches that mix new task data with a small buffer of past task examples~\cite{scialom2022fine,wang2024inscl},
(2) architecture-based methods that introduce new components like adapters to incorporate new tasks~\cite{gururangan2021demix,qin2022elle,zhao2024sapt,wang2024rehearsal},
and (3) parameter-based approaches that either apply regularization to penalize changes in important parameters for old tasks~\cite{zheng2023learn,zhu2024model} or update parameter gradients for each task into orthogonal subspaces~\cite{wang2023orthogonal}. 

However, rehearsal-based methods require data from previously learned tasks, which is not always available~\cite{touvron2023Llama}. 
The architecture and parameter-based approaches typically rely on task labels to design techniques to mitigate gradient conflicts between tasks by updating the parameters task-wise. 
However, obtaining accurate task labels can be challenging or infeasible in LMs' scenarios. 
This paper explores an alternative approach, examining whether the model's inherent features or behaviors can be utilized instead of task labels to mitigate gradient conflicts between tasks. 
Concretely, we examine the distribution of the L1-normalized output magnitude of the linear layers in LMs. 
The output is computed as the dot product between the input $\mathbf{x}\in \mathbb{R}^{d_\text{in}}$ and the weight $\mathbf{W} \in \mathbb{R}^{d_{\text{in}} \times d_{\text{out}}}$ of the layer, and then the output is normalized using the L1-norm, resulting in a vector $\mathbf{n} \in \mathbb{R}^{d_{\text{out}}}$. 
Our analysis reveals an intriguing finding: the L1-normalized output $\mathbf{n}$ exhibits distinct magnitude distributions for different tasks\footnote{The term `(L1-)normalized output magnitude distribution' will be referred to interchangeably as `magnitude distribution' for brevity throughout the paper.}. 
The process and observation described above are illustrated in Figure~\ref{fig:motivation}, which presents real output data for the first linear layer of the Feedforward Network (FFN) in the last Transformer block of T5. We can observe that the magnitude distributions differ significantly for three example tasks - BoolQA, COPA, and Yelp. 
Motivated by this observation, we argue that the differences in magnitude distributions within LMs could serve as a natural, label-free alternative to replace the need for external task labels in mitigating gradient conflicts to update the model's parameters task-wise. However, this potential is locked during conventional continual learning settings. 

To this end, we introduce ``MIGU'' (\textbf{M}agn\textbf{I}tude-based \textbf{G}radient \textbf{U}pdating for continual learning), leveraging the inherent differences in magnitude distributions of the L1-normalized output in LMs' linear layers to enable continual learning without relying on task labels. Specifically, during the forward propagation phase, we cache and normalize the output of the linear layers using the L1-norm. 
Then, in the backward propagation phase, we only update the parameters with the $T$ largest values in L1-normalized magnitude, where $T$ is a predefined threshold ratio. 
Since different tasks exhibit distinct magnitude distribution patterns, MIGU can effectively harness the LMs' inherent features to update the parameters with large magnitudes per task, alleviating gradient conflicts and unlocking their innate continual learning potential.

We evaluate MIGU across three main LM architectures: the encoder-only RoBERTa~\cite{liu2019RoBERTa}, the encoder-decoder T5 model~\cite{raffel2023exploring}, and the decoder-only Llama2~\cite{touvron2023Llama}. 
Furthermore, we consider two continual pre-training settings for LMs: continual pre-training and continual finetuning, using four CL datasets. 
Notably, our approach can seamlessly integrate three mainstream CL approaches - rehearsal-based, architecture-based, and parameter-based - to further enhance the CL abilities of LMs. 
When evaluated on the four datasets, our experimental results achieve comparable or superior performance to the current state-of-the-art methods. 
For example, in a 15-task long sequence CL dataset, the MIGU leads to a 15.2\% accuracy improvement over the conventional parameter-efficient finetuning baseline. 
Furthermore, combining MIGU with three types of CL methodologies substantially improves these individual CL approaches. 
We also provide detailed ablation studies and visualizations on MIGU, revealing that CL with MIGU pushes the magnitude distribution similarity between tasks farther apart and better avoids conflicts. 
We believe the work presents a novel perspective on exploring CL in LMs. 
Our code is publicly available at {https://github.com/wenyudu/MIGU}.

\section{Related Work}\label{sec:related}

\paragraph{Continual Learning for Language Models.}

Continual learning is a long-standing challenge through the history of machine learning and deep learning~\cite{mccloskey1989catastrophic,wu2024continual}. 
Recent studies for CL in LMs can be roughly categorized into three categories. 
1. Rehearsal-based approach that mixes new task data with a small buffer of past task examples~\cite{scialom2022fine,wang2024inscl}. 
2. Architecture-based approach that expands new modules like adapters to incorporate new tasks~\cite{gururangan2021demix,qin2022elle,zhao2024sapt,wang2024rehearsal}. 
3. Parameter-based method that updates parameters in a task-aware manner. 
Some literature~\cite{Wang2023ACS} splits the parameter-based into either regularization-based approaches that add a regularization term to penalize changes in important weights of the earlier learned tasks~\cite{zheng2023learn,zhu2024model}, or optimization-based approaches that updates parameters gradients for each task into orthogonal subspaces to avoid conflicts~\cite{wang2023orthogonal}. 
These methods rely on either old task data or accurate task labels, which are hard or expensive to collect for LMs' continual training. In contrast, MIGU only leverages LMs' innate features for CL. 

\paragraph{Partially Updating Parameters in Continual Learning.}
Among existing CL methods for LMs, our approach and regularization-based~\cite{zheng2023learn,zhu2024model} approach both partially update parameters, but ours fundamentally diverges from the regularization-based methods in motivation and design. 
While they rely on backward gradients to identify and protect important weights for old tasks, we leverage the differences in magnitude distribution across tasks during the feed-forward phase. 
Additionally, our method's ability to freely mask at the sample level sets us apart from their fixed gradient mask approach. Furthermore, our method does not require task labels, enabling it to work in broader scenarios where task labels are unavailable. 
Lastly, the layer output distributions are naturally obtained during the feed-forward phase training, whereas they normally require an additional subset to derive the gradient mask before training on a task.

\paragraph{Finding Important Weights.}
One may classify our method as a broader research cluster centered on finding important weights, a topic that has been extensively explored in continual learning~\cite{zhu2023survey}, model pruning and compression~\cite{frankle2019lottery}, efficient training and inference~\cite{ansell2024scaling}, as well as investigations into activation sparsity~\cite{zhang2023emergent,song2024turbo}, and other related areas. 
However, these works mostly use weight or gradient magnitude to define a fixed size of the important weights. 
A few works on activation sparsity use the sparsity patterns after the activation function for either efficient inference~\cite{zhang2022moefication} or performance improvements~\cite{qiu2024unlocking}. 
None of the above explore the general dot product of weights and layer input. 
The closest work to ours is an unstructured pruning work~\cite{sun2024simple} using the dot product of weight and input, demonstrating a superior method to pure weight-based pruning. 
However, this prior work fails to consider the varying patterns of important weights across different tasks. 
In contrast, our method utilizes the L1-normalized dot product of weight and input as an inherent indicator of importance in CL settings.

\section{Method}

In Table~\ref{model comparisons}, we compare MIGU with common CL methods. Our approach is only one rehearsal-free, task-id-free method that supports both continual pre-training and continual finetuning.

\subsection{Preliminary - Continual Learning Setup}

Continual learning~\cite{Ke2022ContinualLO, wang2023trace,zhao2024sapt} aims to tackle the challenges that arise within the ongoing sequence. Formally, tasks $\left\{\mathcal{T}_1, \ldots, \mathcal{T}_T\right\}$ arrive in sequentially. Each task $\mathcal{T}_t=\left\{\left(x_t^i, y_t^i\right)\right\}_{i=1}^{n_t}$ contains a separate target dataset with the size of $n_t$. For any time step $t$, the model is expected to not only adapt itself to the $t$-th task, but also retain its capabilities across all the previous tasks it has been trained on.
This study explores two distinct CL settings. In the first setting, where only the MIGU method is employed, the task label is unavailable during the training and testing phases. Secondly, when combined with the three existing types of CL techniques, the model can be exposed to old task data or task information during the training phase.

\begin{table}[t]
\begin{center}
\resizebox{1.0\columnwidth}{!}{
\begin{tabular}{lcccc}
    \toprule
    & \textbf{\ RF\ } & \textbf{\ TIFT\ } & \textbf{\ CIT\ } & \textbf{\ CPT\ }\\ 
    \midrule
LFPT5 \cite{qin2021lfpt5}   &    &      & $\checkmark$   &                      \\
EPI \cite{Wang2023EIP}            &   $\checkmark$  &     & $\checkmark$     &     \\
O-LoRA\cite{wang2023orthogonal} & $\checkmark$   &          & $\checkmark$   &    \\ 
MoCL \cite{wang2024rehearsalfree}   &        $\checkmark$  &                             & $\checkmark$                         &                      \\
SAPT \cite{zhao2024sapt}          &   $\checkmark$         &    & $\checkmark$      &                      \\
DAS \cite{ke2023continual}          & $\checkmark$      &                              &                          &   $\checkmark$                   \\
\midrule
\textbf{MIGU} & $\checkmark$ &$\checkmark$& $\checkmark$& $\checkmark$ \\
\bottomrule

\end{tabular}
}
\caption{The comparison between MIGU and other CL methods. Specifically, \textbf{RF} indicates whether the method is rehearsal-free. \textbf{TIFT} indicates whether the method is task-id-free during training. \textbf{CIT} indicates whether the method supports instruction finetuning.\textbf{CPT} indicates whether the method supports continual pre-training.}
\label{model comparisons}
\end{center}
\end{table}

\subsection{MIGU - MagnItude-based Gradient Updating for Continual Learning.}
\label{subsec:MIGU}

\begin{figure}[h]
    \centering
    \includegraphics[width=1\linewidth]{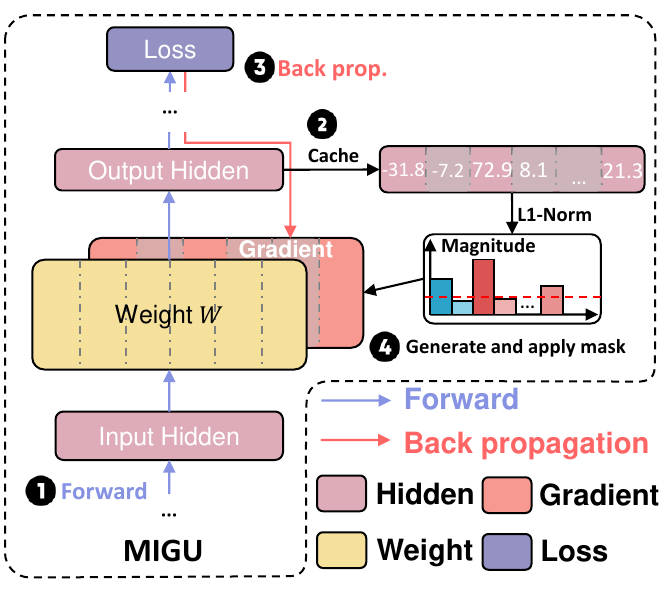}
    \caption{Proposed method: MIGU. During 1) the forward phase, our method 2) caches the output magnitude of the linear layers, and 3) after backpropagation, 4) MIGU masks the gradients by cached magnitudes to update parameters accordingly.}
    \label{fig:method_migu}
\end{figure}

Our approach employs a two-step process to leverage the inherent differences in magnitude distributions across various tasks for continual learning.: 1) Caching output magnitudes and 2) Updating gradient via a magnitude-based mask. We show the process in Figure~\ref{fig:method_migu}. To illustrate our method, we first consider the fundamental component in LMs, a single linear layer\footnote{For simplicity, we omit the bias term $\mathbf{b}$ here.} with weight $\mathbf{W}$
 and only feed an input token $\mathbf{x}$ into LMs.

\paragraph{Feedforward: Caching Output Magnitudes.} 
Given the weight matrix $\mathbf{W} \in \mathbb{R}^{d_{\text{in}} \times d_{\text{out}}}$, we interpret the columns of $\mathbf{W}$ as a set of $d_{\text{out}}$ vectors, each with dimension $d_{\text{in}}$:
\begin{equation}
    \mathbf{W} = [\mathbf{w_1},\dots, \mathbf{w_i},\dots \mathbf{w_{d_{out}}}], \text{where } \mathbf{w_i} \in \mathbb{R}^{d_{\text{in}}}
\end{equation}
Given the input vector of the layer $\mathbf{x} \in \mathbb{R}^{d_{\text{in}}}$, the operation of the layer can be viewed as the dot product between $\mathbf{x}$ and each weight vector $\mathbf{w_i}$: 
\begin{equation}
    {h_{i}} = \mathbf{x} \cdot \mathbf{w_i}
\end{equation}
We then compute the normalized product magnitude ${n_{i}}$ using the L1-norm by : ${n_{i}} = \| {h_{i}} \|_1$, where $\| \cdot \|_1$ denotes the L1-norm. 
Thus, we have the L1-normalized magnitude product distribution vector $\mathbf{n}$ for $\mathbf{W}$.

\paragraph{Backward Propagation: Updating gradient via a magnitude-based mask.}
After calculating the gradient in the backward phase, we obtain the gradient matrix $\nabla\mathbf{W}$ for the weight $\mathbf{W}$, which presents the optimization direction given the input $\mathbf{x}$. 
We then define a mask matrix $\mathbf{M}$ to partially mask $\nabla\mathbf{W}$ using the L1-normalized product magnitudes cached during the forward phase. 
Formally, we sort the product magnitudes in the descending order and mask the corresponding gradients as follows:

\begin{align}
{t} &= \lfloor T \times d_{\text{out}} \rfloor \\
 \mathbf{M} &=\text{BinaryTopT}(\textbf{n}, t)
\end{align}
\begin{equation}
\begin{array}{l}
\text{BinaryTopT}(\mathbf{n_i}, t)\\
=\begin{cases} 
1 & \text{if } \mathbf{n_i} \text{ is in the top } 1-t \text{ elements of } \mathbf{n} \text{.} \\
0 & \text{otherwise,}
\end{cases}
\end{array}
\end{equation}
where $T$ is the threshold ratio to mask gradient,  ${t}$ is the actual number t to mask, $\lfloor . \rfloor$ is the floor rounding. 
The model update rule is then given by:

\begin{equation}
\mathbf{W}_{\text{new}} \leftarrow  \mathbf{W} - \eta \cdot \mathbf{M} \odot \nabla\mathbf{W}
\end{equation}
where $\eta$ is the learning rate. This formulation ensures that only those weights with L1-normalized magnitudes exceeding the threshold ${T}$ are updated.

\subsection{MIGU in Practice}
In practice, to apply MIGU, we average the product magnitudes of all tokens on a batch to generate the mask for simple implementation.

\paragraph{MIGU in Transformer Block.} For a Transformer block, we apply our method from Section~\ref{subsec:MIGU} to the Query, Key, Value, and Output linear layer of the multi-head attention (MHA) component, and two~(for T5 and RoBERTa) or three~(for Llama) linear layers in the FFN component.

\paragraph{MIGU in LoRA Implementation.}

We also implement MIGU for parameter-efficient finetuning (PEFT) of LMs, particularly we employ Low-Rank Adaptation (LoRA)~\cite{hu2022lora}. The standard LoRA is mathematically represented as follows:
\begin{align}
    \mathbf{x_{A}} &= \mathbf{x} \cdot \mathbf{A} \\
    \mathbf{x_{B}} &= \mathbf{x_{A}} \cdot \mathbf{B}\label{eq:xb} \\
    \mathbf{x_O} &= \mathbf{x} \cdot \mathbf{W} + \frac{\alpha}{r} \cdot \mathbf{x_{B}}\label{eq:xo},
\end{align}
where $\mathbf{x}$ denotes the input representation of the layer, $\mathbf{A} \in \mathbb{R}^{d_{in}\times r}$ and $\mathbf{B} \in \mathbb{R}^{r\times d_{out}}$ are the low-rank matrices, $\alpha$ is a scaling constant, $\mathbf{W}$ is the original weight matrix of the standard linear, and $\mathbf{x_O}$ is the output after applying the LoRA transformation.

To implement MIGU, we apply the same method in Section~\ref{subsec:MIGU} for the matrix $\mathbf{A}$. But for the matrix $\mathbf{B}$, we use the output of $\mathbf{x_O}$ in Equation~\ref{eq:xb} rather than the output of $\mathbf{x_B}$ in Equation~\ref{eq:xo} to compute the magnitude distribution vector.

\section{Experiments}

\begin{table}
    \centering
    \resizebox{0.95\linewidth}{!}{
    \begin{tabular}{l|cc}
    \hline
    \toprule
    \textbf{Methods} & \textbf{Standard} & \textbf{Long} \\
    \midrule
    LFPT5$^*$ & 72.7 & 69.2\\
    EPI$^*$ & 65.3 & -\\
    MoCL$^*$ & 75.9 & -\\
    SAPT-LoRA$^*$ &-  & \textbf{82.0} \\
    MTL$^*$ &\textbf{80.0} & \textbf{80.0} \\
    IncLoRA& 68.8& 64,7\\
    \quad + MIGU & 76.4{\small(7.6$\uparrow$)} & 68.7{\small(4.0$\uparrow$)}\\
    OIncLoRA & 75.8 & 69.6\\
    \quad + MIGU & \textbf{76.6}{\small(0.8$\uparrow$)} & 70.0{\small(0.4$\uparrow$)}\\
    LoRAReplay & 74.5 & 75.2\\
    \quad + MIGU & 76.2{\small(1.7$\uparrow$)} & 76.5{\small(1.3$\uparrow$)}\\
     \midrule
     MoELora &54.1  & 27.6 \\
    FT & 75.7 & 68.3 \\
    \quad + MIGU & \textbf{78.8}{\small(3.1$\uparrow$)} & \textbf{73.8}{\small(5.5$\uparrow$)}\\
    LoRA & 67.9 & 46.0 \\
    \quad + MIGU & 73.3{\small(5.4$\uparrow$)} & 61.2{\small(15.2$\uparrow$)}\\
    \bottomrule    \end{tabular}}
    \caption{Average accuracy on standard CL benchmark (Order 1,2,3) and long CL benchmark (Order 4,5,6) with T5-large model. The top block contains CL methods with extra old task data or task labels, while the bottom does not. Methods denoted with $^*$ are copied from previous papers \cite{wang2023orthogonal,wang2024rehearsalfree,zhao2024sapt}.}
    \label{tab:my_label_t5}
\end{table}

We use three language models adopted by the previous lines of works in CL for NLP: encoder-only RoBERTa~\cite{liu2019RoBERTa}, encoder-decoder T5 model~\cite{raffel2023exploring} and decoder-only Llama2~\cite{touvron2023Llama}. 
We start with continual finetuning T5-large \cite{raffel2020exploring} on two CL datasets following the settings from~\cite{qin2021lfpt5,wang2023orthogonal}.
We implement MIGU upon vanilla finetuning and PEFT with LoRA~\cite{hu2022lora}. 
We also combine our method with three main types of CL approaches to examine the seamless integration of our method with the existing CL methodologies. 
Next, we use encoder-only RoBERTa to continual pre-traning domain adaptive data, following the setting~\cite{ke2023continual}. 
We further scale our experiment to decoder-only Llama2-7B~\cite{touvron2023Llama} and test the trade-off between base model ability and new task ability.
All experimental results are reported as the average of 3 runs.
Please refer to the Appendix \ref{T5_detailed_setting} for more detailed settings.

\subsection{Continual Finetuning on T5-large}\label{CIT_T5}
\paragraph{Two Benchmarks.}
We evaluate our approach to continual finetuning on T5-large using the standard CL benchmark and long sequence benchmark. 
We follow the setup from \cite{qin2021lfpt5,wang2023orthogonal} to shuffle the four text classification tasks from the LM dataset ~\cite{zhang2015character} into three different orders to form Order 1, 2, 3 for standard CL benchmark. Similarly, we shuffle a mix of 15 tasks (five classification tasks, nine GLUE and SuperGLUE tasks, and the IMDB dataset) to form Orders 4, 5, and 6 for the long sequence benchmark. 
For the details on benchmark and sequence, please refer to the appendix~\ref{baseline_t5_large}.

\paragraph{Baselines.} 
We separate the baselines into two categories: without old data or task information  and with old data or task information during training. 
For the first category, we include vanilla \textbf{FT}, which trains all model parameters on a sequence of tasks, and vanilla \textbf{LoRA}, in which fixed-size LoRA parameters are trained on a sequence of tasks. For the second category, we have rehearsal-based approaches: \textbf{LoRAReplay} that trains new tasks on LoRA with mixing a 2\% past task, \textbf{LFPT5} \cite{qin2021lfpt5} continuously trains a soft prompt that simultaneously learns to solve the tasks and generate training samples for experience replay; architecture-based approaches: \textbf{IncLoRA} that incremental learning of new LoRA parameters on a sequential series of tasks, \textbf{MoELora}~\cite{luo2024moelora}, a vanilla MoE with LoRA number equals to the task number, \textbf{SAPT-LoRA}~\cite{zhao2024sapt} extends IncLoRA by aligning learning process and selection process of LoRA, and \textbf{MoCL}~\cite{wang2024rehearsalfree} continually adds new modules and composes them with existing modules; 
parameter-based approaches \textbf{OIncLoRA}~\cite{wang2023orthogonal}\footnote{O-LoRA is original name, we rename it to OIncLoRA to emphasize it is build upon IncLoRA and align with our notation.} extends IncLoRA to learn different LoRAs into orthogonal subspaces. 
Moreover, we have one multi-task learning baseline \textbf{MTL} from~\cite{wang2023orthogonal} as the referenced ``upper bound'' for the benchmark.

\paragraph{Metrics.}
ACC (Accuracy~\cite{Chaudhry_2018}). The average performance of all tasks after training on the last task, i.e., $A_{\mathcal{T}}=\frac{1}{\mathcal{T}} \sum_{t=1}^{\mathcal{T}} a_{\mathcal{T}, t}$.

\paragraph{Results on T5.}

Table~\ref{tab:my_label_t5} shows that our proposed approach (+MIGU) improves the performance of all five CL approaches. 
Notably, when our method is applied, the vanilla FT and LoRA baselines see substantial improvements. 
Some results obtained using our approach are comparable to the SOTA CL methods that leverage task labels or old task data. 
Notably, the LoRA+MIGU approach surpasses the vanilla LoRA method by a substantial 15.2\% on the long sequence benchmark, significantly mitigating the drawbacks of LoRA in the CL setting with long sequences. 
We choose to combine our method with three LoRA-based techniques to integrate with three CL approaches that leverage old data or additional labels. 
The parameter-based IncLoRA+MIGU exhibits the most significant improvement over the original IncLoRA, implying that our magnitude-based approach can effectively mitigate the conflicts among the sequentially learned LoRA parameters in IncLoRA. 
The relatively marginal improvement of parameter-based OIncLoRA+MIGU indicates a similar function between our approach and projecting LoRAs into orthogonal subspaces, but our method does not require task labels during the continual training process. 
SAPT-LoRA achieves the SoTA performance in long sequence benchmark, but it requires both task labels and past data, which are often infeasible or costly in LMs settings. 
We also report an efficiency study in Appendix~\ref{das_detailed} Table~\ref{tab:efficiency} to show our approach only leads to a minor overhead over the vanilla methods, which is assumed to be more efficient than other CL methods.
We provide a full experiment in Appendix~\ref{Experiment_on_T5} Table \ref{detained_main_results}. 
We also draw the Violin Plot to show the statistical significance of our approach over baselines in Appendix \ref{Experiment_on_T5}.

\subsection{Continual Pre-training on RoBERTa}

\paragraph{Benchmark.} 
In contrast to the previous continual finetuning setting, \citet{ke2023continual} introduces DAS, a new benchmark for continual pre-training (CPT) of LMs. 
DAS is composed of six unlabeled domain corpora, which contain three review domains and three academic paper domains. It is then evaluated using six corresponding classification datasets. 
Unlike continual finetuning, CPT is carried out in two stages: 1) continual sequential pre-training on each domain, and 2) separately continual finetuning for end tasks in each domain.\footnote{Let's take DAS as an example. Stage 1: sequentially continual pre-train Roberta on domains 1-6; stage 2: duplicate Roberta after stage 1 into six copies and continual finetune each copy on separate domain-relevant task from 1 to 6 respectively.} 
Please refer to the Appendix \ref{dataset_instruction} for the details. 

\paragraph{Metrics.} For continual pre-training, we utilize MF1 (Macro-F1) and ACC (Accuracy) following \cite{ke2023continual} to evaluate the performance after pre-training on the last domain. For the details, please refer to \cite{ke2023continual}.

\paragraph{Baselines.} We choose top baselines ranging from vanilla methods that pre-train RoBERTa on domains sequentially with full parameters \textbf{FT} and with PEFT \textbf{Adapter} to rehearsal-based (\textbf{DER++}~\citep{buzzega2020dark}), architecture-based (\textbf{DEMIX}~\citep{gururangan2021demix}), and parameter-based \textbf{HAT-Adapter}~\citep{Serra2018overcoming} and \textbf{DAS}~\cite{ke2023continual}.
\paragraph{Results on RoBERTa.}
\begin{table}
    \centering
    \resizebox{1.0\linewidth}{!}{
    \begin{tabular}{l|cc}
    \toprule
    \textbf{Methods} & \textbf{MF1} & \textbf{ACC} \\
    \midrule
    DEMIX$^*$ & 74.70 & 79.66 \\
    DER++$^*$ & 75.78 &  80.46\\
    HAT-Adapter$^*$ & 74.63 & 79.78 \\
    DAS$^*$ & \textbf{77.90} & \textbf{81.90} \\
    DAS & 76.59 & 81.07\\	
    \midrule
    Adapter & 74.05 &79.48 \\
    FT & 76.36 & 80.77  \\
    \quad + MIGU & \textbf{76.73}{\small(0.37$\uparrow$)} & \textbf{81.19}{\small(0.42$\uparrow$)} \\ 
    \bottomrule
    \end{tabular}
    }
    \caption{Average MF1, ACC on the DAS benchmark after continual pre-training on all domains and finetuning on their corresponding end-task datasets.  The top block contains CL methods with extra old task data or task labels, while the bottom does not. Methods denoted with $^*$ are copied from original papers.}
    \label{tab:improved_style_with_category_and_multirow}
\end{table}
\begin{table}
    \centering
    \begin{tabular}{l|cc}
    \hline
    \toprule
     & \textbf{Avg. Domain 1-2} & \textbf{Avg. Domain 5-6} \\
    \midrule
    FT & 79.69 & 82.07\\
    DAS & \textbf{80.30}\small{(0.61\textcolor{red}{$\uparrow$})} & 81.16\small{(0.91\textcolor{blue}{$\mathbf{\downarrow}$})}\\
    MIGU & 80.14\small{(0.45\textcolor{red}{$\uparrow$})} & \textbf{82.41}\small{(0.34\textcolor{red}{$\uparrow$})}\\
    \bottomrule
    \end{tabular}
    \caption{The average ACC of the first and last two learned domains in the DAS benchmark.}
    \label{tab:verage_two_dataset}
\end{table}

We evaluate MIGU in another setting in which, we continually pre-train a RoBERTa model to six domains sequentially (domain-adaptive pre-training).
Our experimental results in Table \ref{tab:improved_style_with_category_and_multirow} also show promising results of our approach over or on par with the sophisticated CL methods with task labels or old data.
For instance, FT+MIGU achieves 0.37\% improvement in MF1 and 0.42\% in ACC.
We also explore the performance of the domains in different orders. 
We report the average ACC of the first and last two learned domains in Table \ref{tab:verage_two_dataset}.
The results indicate that while the DAS model exhibits less forgetting in the earlier learned domains, but it also learns less in the last domains, possibly due to the strong regularization used to constrain its parameter updates during the CL process over a long sequence. In contrast, MIGU demonstrates a more sustainable method, exhibiting robust performance on the earlier and recently learned domains.

\subsection{Forgetting Less and Learning the Same: Scaling to Llama2}

\paragraph{Results on Llama2.}
\begin{figure}[h]
    \begin{minipage}{1\linewidth}
       \subfigure[Learn the same. Instruction tuning results on Human eval. MIGU with LoRA learns the same as the valinna LoRA.]{
        \includegraphics[width=\linewidth]{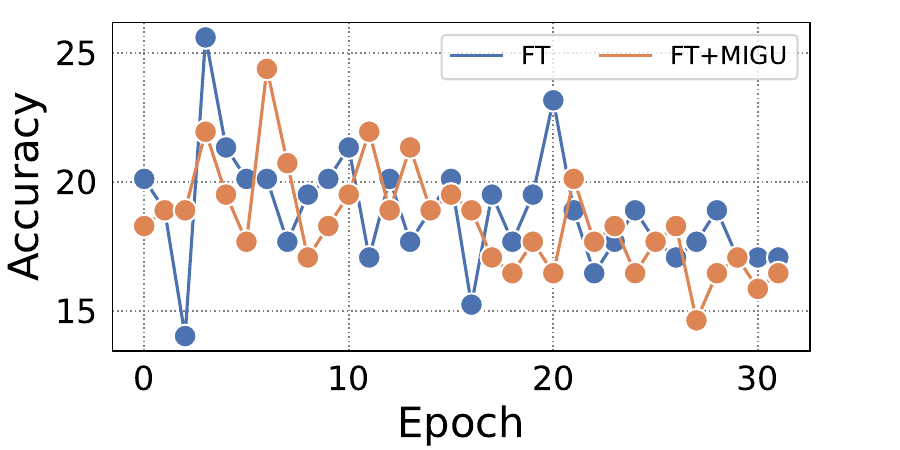}
        }
    \end{minipage}
    \centering
        \begin{minipage}{1\linewidth}
       \subfigure[Forget less. Average accuracy on HellaSwag, Winogrande, ARC-Challenge for Llama-2-7B. The results indicate that MIGU with LoRA forgets less than valinna LoRA.]{
        \includegraphics[width=\linewidth]{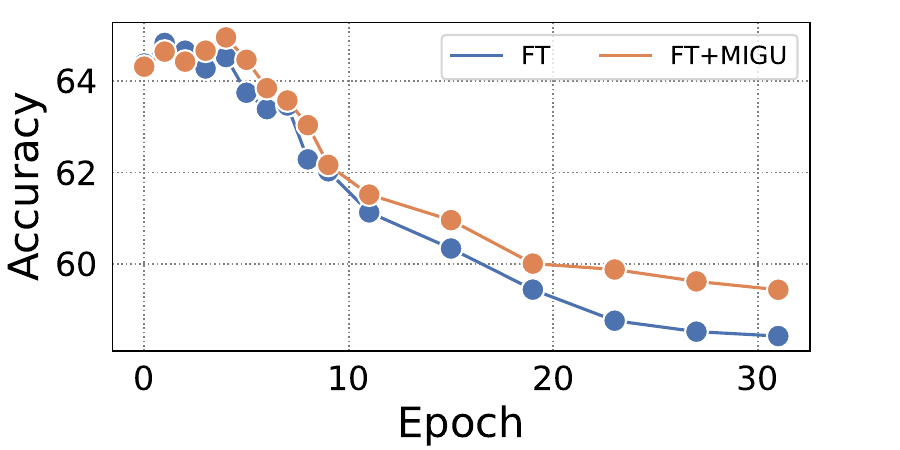}
        }
    \end{minipage}
    \caption{Performance comparison of LoRA with MIGU and the baseline vanilla LoRA on Llama2-7B instruction tuning, evaluated using the Humaneval~\cite{chen2021evaluating}, as well as on LM benchmarks: HellaSwag~\cite{zellers2019hellaswag}, Winogrande~\cite{sakaguchi2019winogrande}, and ARC-Challenge~\cite{clark2018think}.}
    \label{fig:Llama_code_full}
\end{figure}
We further assess our approach on a more demanding LLM continual instruction tuning setting. We finetune a base Llama2-7B on Magicoder-Evol-Instruct-110K for 32 epochs. 
This dataset~\cite{wei2024magicoder} contains 72.97M tokens of programming questions and answers. However, due to computation constraints, we sample 20\% of data and conduct experiments on LoRA. 
We follow~\cite{biderman2024lora} to assess LoRA+MIGU's capabilities on both the base ability (forgetting domain) and the code ability (learning domain). 
To evaluate code learning performance, we utilize the Humaneval benchmark~\cite{chen2021evaluating}, which contains 164 problems that generate a Python program with a docstring and a function signature. A generation is considered correct if it passes all supplied unit tests. 
To quantify how much they have forgotten previous knowledge, we follow~\cite{biderman2024lora} that utilizes average scores of three benchmarks, HellaSwag~\cite{zellers2019hellaswag}, WinoGrade~\cite{sakaguchi2019winogrande} and ARC-challenge~\cite{clark2018think}. 
The experiments are shown in Figure \ref{fig:Llama_code_full}. Compared to baseline FT, our method learns a similar level of new code knowledge but exhibits significantly less forgetting of previous knowledge. This suggests our approach achieves a better trade-off point on the Pareto frontier between learning plasticity and memory stability~\cite{1189626,wang2024comprehensive}. 
For example, after 32 training epochs, the average accuracy across the three benchmarks for our method is 59.4, while the baseline model only achieves 58.4.

\section{Discussions}
We then provide ablations on gradient mask threshold and components as well as a visualization. We also ablate the various methods for finding important weights in Table~\ref{tab:abl_gradient_importance_method} of Appendix.

\subsection{Ablation on Gradient Mask Threshold}

\begin{figure}[h]
    \centering
    \includegraphics[width=1\linewidth]{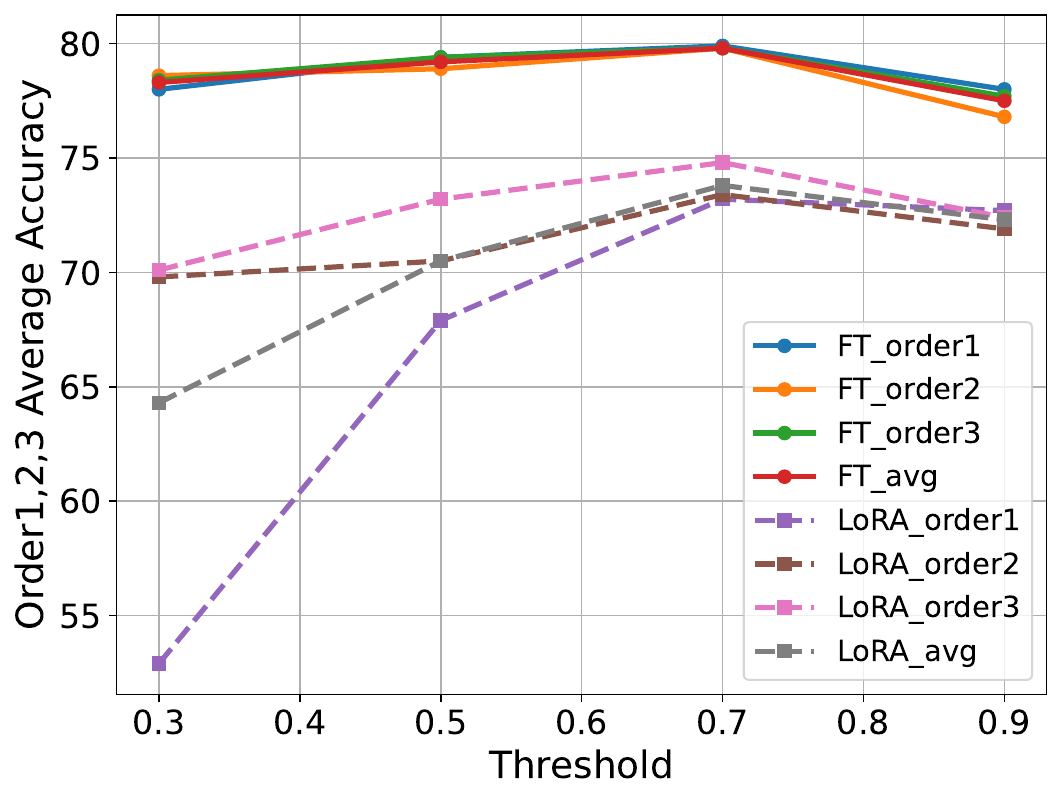}
    \caption{Ablation study on the gradient mask threshold. The curves illustrate that the optimal value is concentrated around 0.7 for FT+MIGU and LoRA+MIGU.}
    \label{fig:threshold_comparison}
\end{figure}

Because the current instruction finetuning datasets like standard CL benchmarks, typically lack a separate development set, we perform an ablation study to manually create a development set of 1,000 samples for each task from the training data in the standard CL benchmark. We use this new split dataset to evaluate the mask threshold selections for the FT, FT+MIGU, LoRA, and LoRA+MIGU methods. Our findings are presented in Table~\ref{fig:threshold_comparison}, indicating that a threshold of 0.7 is optimal for both FT+MIGU and LoRA+MIGU. Additional experiments on mask thresholds for various methods can be found in Appendix~\ref{more_threshold}.

\subsection{Ablation on Gradient Mask Components}

\begin{table}
    \centering
    \resizebox{1.0\linewidth}{!}{
    \begin{tabular}{lcccc}
    \toprule
     & \textbf{Order1} & \textbf{Order2} & \textbf{Order3} & \textbf{Avg} \\
    \midrule
    FT + MIGU & 79.6 & \textbf{80.3} & \textbf{79.2} & \textbf{79.7} \\
    \midrule
    FT & 75.3 & 76.1 & 78.0 & 76.5 \\
     + FFN $1_{st}$-L & 76.9 & 75.9 & 75.8 & 76.2 \\
     + FFN all & 77.2 & 77.2 & 76.7 & 77.0 \\
     + Attention Q   &  77.2& 76.4  & 78.3 & 77.3  \\
     + Attention K   & 76.9 & 73.4  & 75.6 & 75.3  \\
     + Attention V   & 75.4 & 76.3  & 78.0 & 76.6 \\
     + Attention O   & 76.2 & 75.9  & 76.0 & 76.0 \\
     + Attention all  & \textbf{80.2} & 78.8  & 79.0 &  {79.3} \\
    \bottomrule
    \end{tabular}}
    \caption{The ablation results from applying MIGU to different LM components.``+ FFN all'' means only applying MIGU to all the linear operators in FFN layers. The results demonstrate implementing MIGU across all linear layers leads to the most benefits.}
    \label{tab:abl_components}
\end{table}

We further investigate which components within a transformer block should utilize MIGU. 
Typically, a transformer block consists of six linear layers: the query, key, and value (QKV) linear layers and the output linear layer (O) in the MHA module, as well as the two linear layers in the FFN. Our analysis in Table~\ref{tab:abl_components} shows that employing MIGU across all these linear layers achieves the best overall performance, suggesting that the magnitude-based approach is effective for linear layers in different parts of the transformer architecture.

\subsection{Visualization}
\begin{figure}[h]
    \centering
    \begin{minipage}{1\linewidth}
        \subfigure{
        \includegraphics[width=0.49\linewidth]{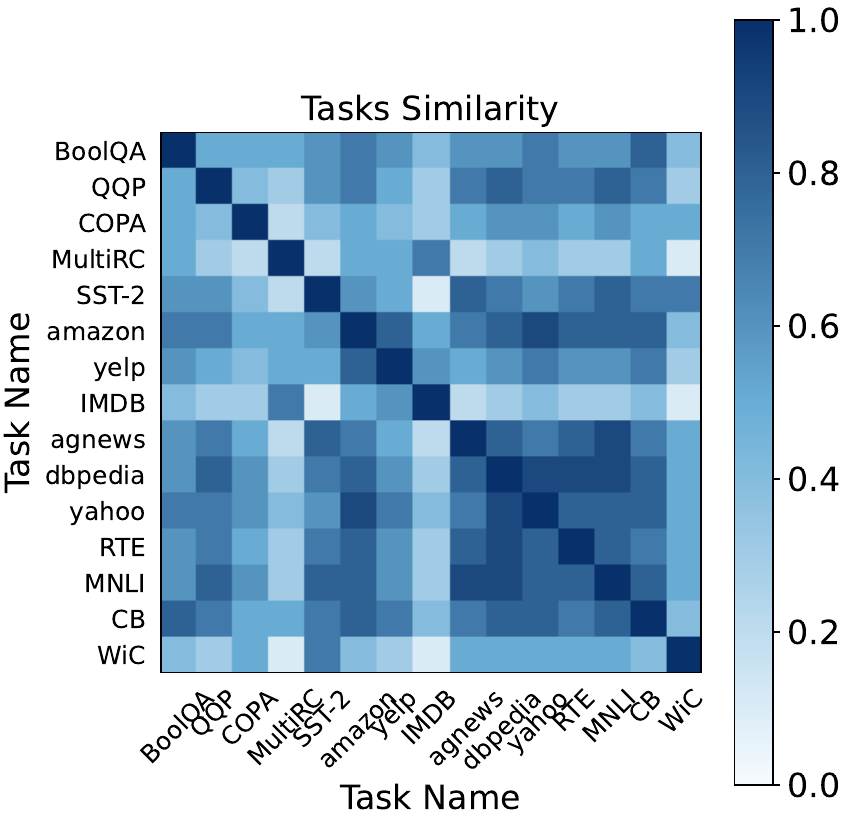}
        }\noindent
         \subfigure{
        \includegraphics[width=0.49\linewidth]{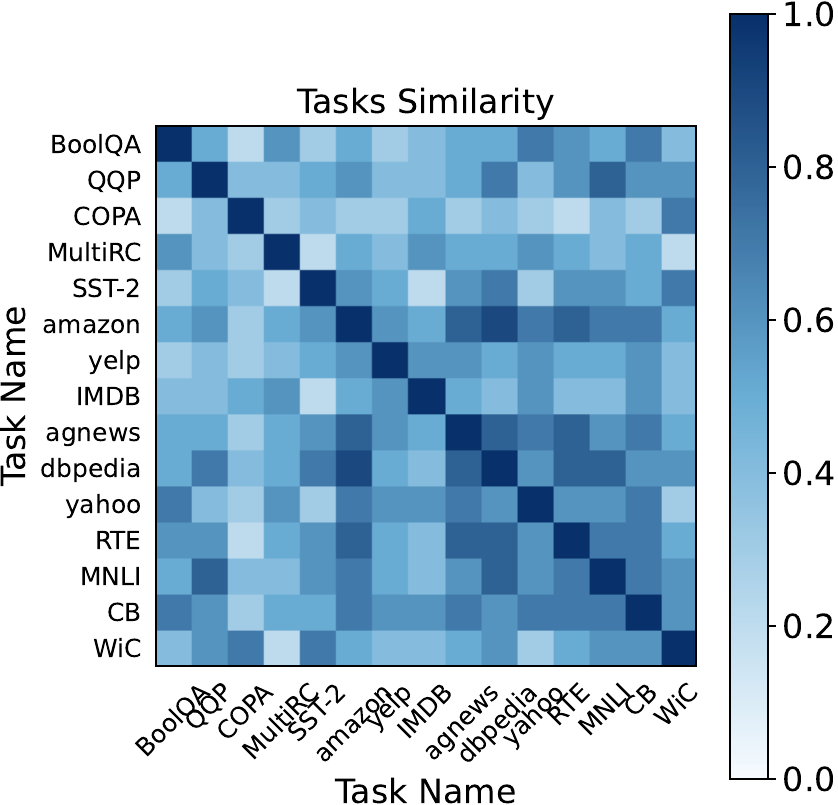}
        }
    \end{minipage}
     \setcounter{subfigure}{0}
    \begin{minipage}{1\linewidth}
        \subfigure[FT]{
        \includegraphics[width=0.49\linewidth]{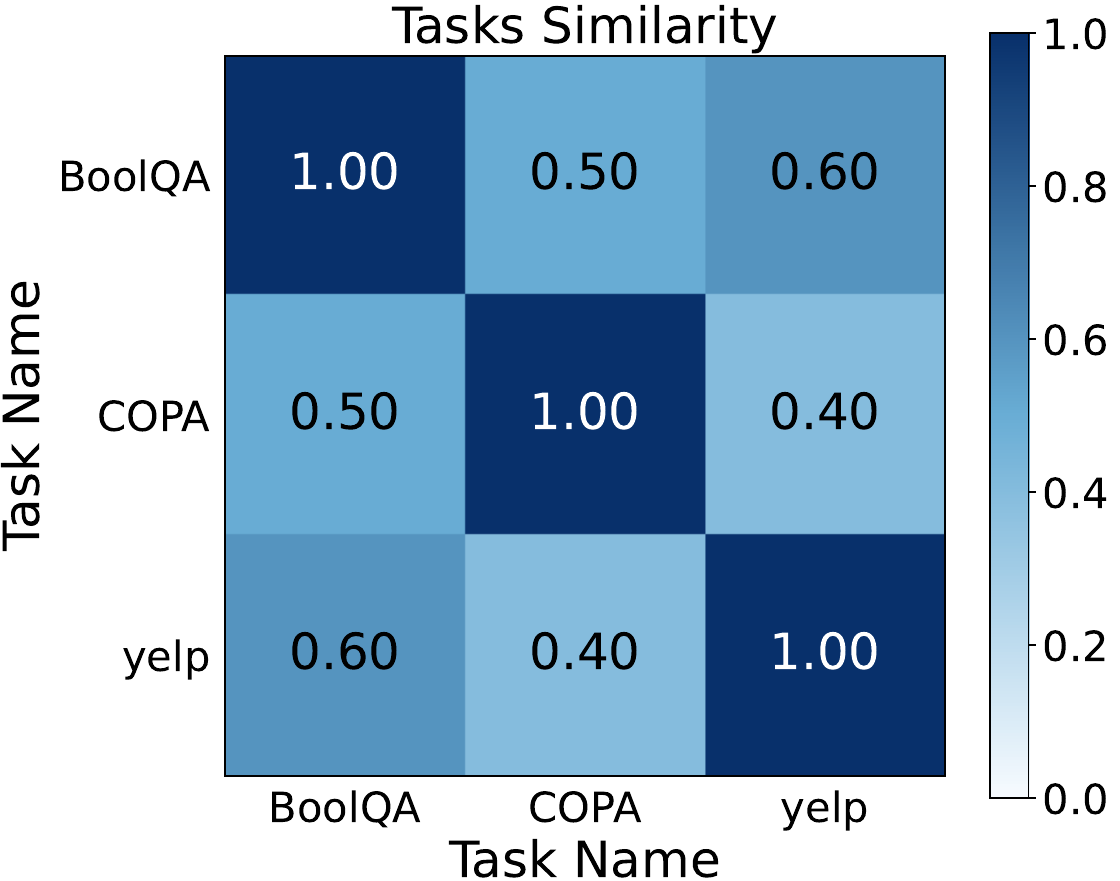}
        }\noindent
         \subfigure[FT + MIGU]{
        \includegraphics[width=0.49\linewidth]{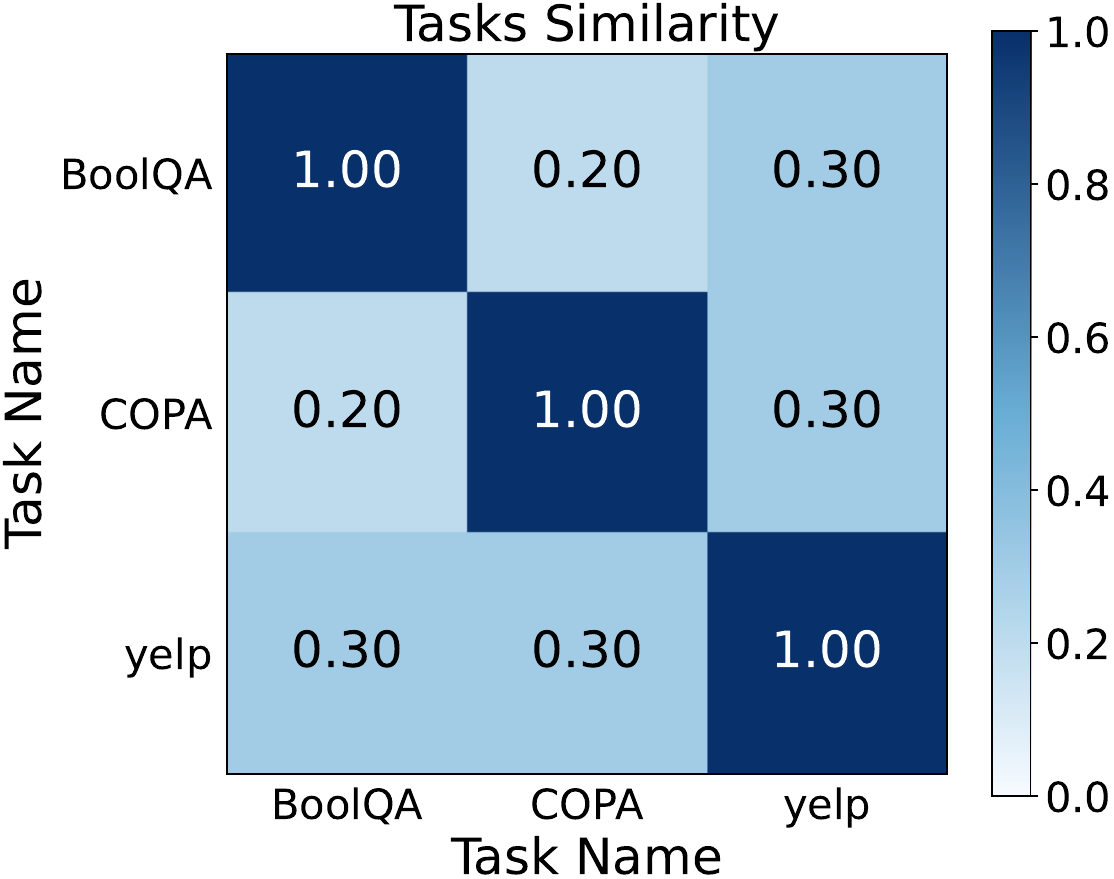}
        }
    \end{minipage}
    \caption{The visualization of magnitude distribution similarity across different tasks.  FT+MIGU is lower, indicating that MIGU reduces the possibility of weight conflicts between tasks. The two sub-figures at the bottom are three highlighted task samples: BoolQA, COPA and Yelp.}
    \label{fig:visualization_figure5}
\end{figure}

We evaluate task similarity by counting the overlapping ratio of updated parameters (large magnitudes) positions by using 100 samples per task.
In Figure \ref{fig:visualization_figure5}, we visualize the task similarity for the first layer of FFN in the last Transformer block of T5-large, comparing FT and FT+MIGU in the Order 6 setting. 
The results clearly show that MIGU increases the degree of parameter isolation across tasks, achieving a similar effect by using task information but without relying on such explicit task labels. We further highlight the similarity between the BoolQA, COPA, and Yelp tasks and the notable decrease in similarity among these three tasks. 
Analyzing the performance results shown in Table \ref{tab:single_dataset}, we find that the significant reduction in overlapping ratio across tasks considerably alleviates the task conflicts, resulting in much more significant performance gains. For example, the accuracy improvement for the COPA dataset is exactly 10\%.
We put the full visualization of all linear layers in Appendix~\ref{visualization}. 

\begin{table}
    \centering
    \resizebox{1.0\linewidth}{!}{
    \begin{tabular}{lccc}
    \toprule
     & \textbf{BoolQA} & \textbf{COPA} & \textbf{Yelp} \\
    \midrule
    FT    &  67.3 &  45.0 & 39.1 \\
    \midrule
     FT + MIGU &  78.3\small{(11.0$\uparrow$)} & 55.0\small{(10.0$\uparrow$)} &47.6\small{(8.5$\uparrow$)} \\
    \bottomrule
    \end{tabular}}
    \caption{The improvement on BoolQA, COPA and Yelp in Order 6.}
    \label{tab:single_dataset}
\end{table}

\section{Conclusion}
We propose MIGU, a rehearsal-free and task-label-free method that only updates the model parameters with large output magnitudes in LM's linear layers. 
By imposing this simple constraint on the gradient update process, we can leverage the inherent behaviors of LMs, thereby unlocking their innate CL abilities. 
Our experiments, applied to all three LM architectures (T5, RoBERTa and Llama2), on two CL scenarios (continual finetuning and continual pre-training) and four CL benchmarks, consistently deliver better performance. Our method can also be seamlessly integrated with existing CL solutions to further improve their performance. 

\section{Limitations}
We acknowledge two limitations for this work.
Due to computation limitations, although we finetune Llama2-7B with LoRA, we are unable to scale our experiments to LM continual pre-training or full tuning. However, our experimental performance on continual pre-training RoBERTa indicates the great potential for the scalability of this general approach.
Another limitation is we only explore an approach for unlocking the inherent CL potential of LMs through updating the gradient by the magnitude of output. There exists more discussions on exploiting innate features such as activation sparsity as discussed in the Related Work section. Though we include our preliminary results in Table~\ref{tab:abl_gradient_importance_method} of Appendix, the limitation can be further addressed in future work.

\section{Acknowledgment}

We thank all constructive comments from anonymous reviewers. 
Reynold Cheng and Wenyu Du were supported by the Hong Kong Jockey Club Charities Trust (Project 260920140), the University of Hong Kong (Project 109000579), the HKU Outstanding Research Student Supervisor Award 2022-23, and the HKU Faculty Exchange Award 2024 (Faculty of Engineering).

\bibliography{anthology,acl_latex}

\appendix

\section{Experimental Details}

\label{app:details}

All experiments are run on an A100 $\times$ 8 DGX-machine.

\subsection{Continual finetuning on T5}\label{T5_detailed_setting}
We adapted the code-base from O-LORA.\footnote{https://github.com/cmnfriend/O-LoRA}

\paragraph{Finetuning (FT) and FT with MIGU.}
\begin{itemize}
    \item The batch size is set to 64.
    \item The optimization is performed using the AdamW algorithm with hyperparameters $\beta_1=0.9$, $\beta_2=0.999$, and a weight decay coefficient of $0.01$.
    \item The initial learning rate is set to $1 \times 10^{-4}$, alongside a static learning rate scheduler.
    \item The threshold for mask selection is set at $0.7$ across orders 1 to 6 in the FT+MIGU configuration.
\end{itemize}

\paragraph{Low-Rank Adaptation (LoRA) and LoRA with MIGU.}
\begin{itemize}
    \item LoRA configuration: $r=8, \alpha=32$, dropout $=0.05$.
    \item The learning rate is set to $1 \times 10^{-3}$, with all other hyperparameters being consistent with the FT+MIGU configuration.
\end{itemize}

\paragraph{Incremental LoRA (IncLoRA) and IncLoRA with MIGU.}
\begin{itemize}
    \item For each LoRA module: $r=8, \alpha=32$, dropout $=0.05$.
    \item Hyperparameters are identical to those specified in the LoRA and LoRA with MIGU settings.
\end{itemize}

\paragraph{Order-Incremental LoRA (OIncLoRA) and OIncLoRA with MIGU.}
\begin{itemize}
    \item The threshold for mask selection is set at $0.05$ across orders 1 to 6 in the FT+MIGU configuration.
    \item All remaining hyperparameters are consistent with the LoRA and LoRA with MIGU settings.
\end{itemize}

\paragraph{LoRA Replay and LoRA Replay with MIGU.}
\begin{itemize}
    \item The threshold for mask selection is set at $0.4$ across orders 1 to 6 in the FT+MIGU configuration.
    \item All remaining hyperparameters are consistent with the LoRA and LoRA with MIGU settings.
\end{itemize}

\subsection{Continual pre-training finetune on RoBERTa}
We adapted the code-base from DAS.\footnote{https://github.com/UIC-Liu-Lab/ContinualLM}

\paragraph{Pre-training.}
\begin{itemize}
    \item The batch size is set to 248.
    \item The optimization is performed using the AdamW algorithm with hyperparameters $\beta_1=0.9$, $\beta_2=0.999$, and a weight decay coefficient of $0$.
    \item The initial learning rate is set to $1 \times 10^{-4}$, alongside a linear learning rate scheduler.
    \item The threshold for mask selection is set at $0.7$ on the sequence of tasks.
\end{itemize}
\paragraph{Tuning.}
\begin{itemize}
    \item The batch size is set to 16.
    \item The optimization is performed using the AdamW algorithm with hyperparameters $\beta_1=0.9$, $\beta_2=0.999$, and a weight decay coefficient of $0.01$.
    \item The initial learning rate is set to $3 \times 10^{-5}$, alongside a linear learning rate scheduler.
\end{itemize}
\subsection{Instruct finetuning on Llama2.}
\begin{itemize}
    \item The optimization is performed using the AdamW algorithm with hyperparameters $\beta_1=0.9$, $\beta_2=0.95$, and a weight decay coefficient of $0$.
    \item The initial learning rate is set to $5 \times 10^{-4}$, alongside a cosine learning rate scheduler with warmup $=0.1$ of the total duration.
    \item LoRA configuration: $\alpha=32$, dropout $=0.05$.
\end{itemize}

\section{Benchmark Instruction}
\begin{table*}[h]
\begin{tabular}{lll}
\hline
\textbf{Benchmark} &\textbf{Order}  & \textbf{Task Sequence}\\ \hline
\multirow{3}{*}{Standard CL} & 1    & dbpedia → amazon → yahoo → ag    \\
& 2                 & dbpedia → amazon → ag → yahoo          \\
& 3                  & yahoo → amazon → ag → dbpedia        \\ \hline
\multirow{3}{*}{Long sequence} & 4        & \begin{tabular}[c]{@{}l@{}}mnli → cb → wic → copa → qqp → boolqa → rte → imdb →\\ Yelp → amazon → sst-2 → dbpedia → ag → multirc → yahoo\end{tabular} \\
& 5       & \begin{tabular}[c]{@{}l@{}}multirc → boolqa → wic → mnli → cb → copa → qqp → rte\\ → imdb → sst-2 → dbpedia → ag → Yelp → amazon → yahoo\end{tabular} \\
& 6       & \begin{tabular}[c]{@{}l@{}}Yelp → amazon → mnli → cb → copa → qqp → rte → imdb →\\ sst-2 → dbpedia → ag → yahoo → multirc → boolqa → wic\end{tabular} \\ \hline
DAS  & 7 & Restaurant $\to$ ACL $\to$ AI $\to$ Phone $\to$ PubMed $\to$ Camera \\ \hline
\end{tabular}
\caption{Task Sequence Orders for Continual Learning Experiments. Orders 1-3 represent the conventional task sequences employed in standard continual learning benchmarks~\cite{zhang2015character}. Orders 4-6 extend to longer sequences, encompassing 15 tasks each~\cite{razdaibiedina2023progressive}. Order 7 comprises a sequence of 6 tasks derived from unsupervised pre-training domains, in accordance with~\cite{ke2023continual}.}\label{table:Training_order}
\end{table*}
\subsection{Dataset Information}\label{dataset_instruction}
Our experimental section encompasses datasets including the \textbf{Standard CL benchmark} and \textbf{Long sequence benchmark}, both of which are utilized for instruction finetuning on the T5-large model; the \textbf{DAS benchmark}, which is used for continual pre-training on RoBERTa; the \textbf{Magicoder-Evol-Instruct-110K}, which pertains to instruction tuning on Llama-2-7B; and the datasets \textbf{Hellaswag}, \textbf{WinoGrande}, and \textbf{ARC-Challenge} for evaluating the finetuned Llama-2-7B.

\paragraph{Standard CL benchmark.}\label{Standard} For continual finetuning, we use MTL5 dataset introduced by~\cite{zhang2015character}, and follow the setup from LFPT5 and O-LoRA \cite{qin2021lfpt5,wang2023orthogonal} to pick four text classification datasets (AG News, Amazon reviews, DBpedia and Yahoo Answers) and shuffle the tasks into three different orders.

\paragraph{Long sequence benchmark.} 
\cite{razdaibiedina2023progressive} extends the Standard CL benchmark by introducing a long sequence benchmark for continual learning benchmark with 15 datasets. This includes five tasks from CL benchmark, four from GLUE benchmark (MNLI, QQP, RTE, SST2)~\cite{wang2018glue}, five from SuperGLUE benchmark (WiC, CB, COPA, MultiRC, BoolQ)~\cite{wang2018glue}, and the IMDB movie reviews dataset~\cite{maas-etal-2011-learning}. Following~\cite{razdaibiedina2023progressive}, we select 1000 random samples for training each task and hold out 500 samples per class for validation.

\paragraph{DAS Benchmark.} \citep{ke2023continual} introduce a new benchmark for continual pre-training of LMs, which is more challenging as the data required to pre-train is much larger and LMs are easier to forget previous knowledge. 
DAS is composed of 6 unlabeled domain corpora, which contain 3 reviews: Yelp Restaurant
~\cite{DBLP:conf/naacl/XuLSY19}, Amazon Phone~\cite{DBLP:conf/emnlp/NiLM19}, Amazon Camera~\cite{DBLP:conf/emnlp/NiLM19}; 3 of them are academic papers: ACL Papers~\cite{DBLP:conf/acl/LoWNKW20}, AI Papers~\cite{DBLP:conf/acl/LoWNKW20}, and PubMed Papers~\footnote{\url{https://pubmed.ncbi.nlm.nih.gov/}}. and evaluated by 6 corresponding classification datasets are: Restaurant~\footnote{\url{https://alt.qcri.org/semeval2014/task4/}}, Phone\cite{ding2008holistic,hu2004mining}, Camera~\cite{ding2008holistic,hu2004mining}, ACL (ACL-ARC in~ \citep{DBLP:journals/tacl/JurgensKHMJ18}), AI (SCIERC in~\citep{DBLP:conf/emnlp/LuanHOH18}), and PubMed (CHEMPORT in~\citep{kringelum2016chemprot}).

\paragraph{Magicoder-Evol-Instruct-110K.}
This dataset~\cite{wei2024magicoder} contains 72.97M tokens
of programming questions and answers. It reproduces the ``Evol-Instruct'' dataset of WizardCoder~\cite{luo2023wizardcoder}: an LLM (GPT-4) is iteratively prompted to increase the difficulty of a set of question-answer pairs (from Code Alpaca~\cite{codealpaca}). Due to computation constraints, we pick contain 20\% the samples to instruct tuning the Llama-2-7B model.
\paragraph{HellaSwag, WinoGrade and ARC-challenge.}
For how much they forget the old knowledge, we follow the~\cite{biderman2024lora} that averages three benchmarks, HellaSwag~\cite{zellers2019hellaswag}, WinoGrade~\cite{sakaguchi2019winogrande} and ARC-challenge~\cite{clark2018think}. HellaSwag benchmark includes 70K problems, each describing an event with multiple possible continuations. The task is to pick the most plausible continuation, requiring inferences about nuanced everyday situations. WinoGrande benchmark also assesses commonsense reasoning. It includes 44K problems with sentences that require ambiguous pronoun resolution. ARC-Challenge benchmark consists of 7,787 grade-school level, multiple-choice science questions, testing capabilities in complex reasoning and understanding scientific concepts.

\begin{table*}[th]
\centering
\setlength{\tabcolsep}{4pt}{
\begin{tabular}{l|cccc|cccc}
\specialrule{2pt}{2pt}{\aboverulesep}
\multirow{2}{*}{\textbf{Method}}       & \multicolumn{4}{c|}{\cellcolor[HTML]{FFFFFF}\textbf{Standard CL Benchmark (4 tasks)}} & \multicolumn{4}{c}{\cellcolor[HTML]{FFFFFF}\textbf{Longer CL Benchmark (15 tasks)}} \\
        & \textbf{Order-1}       & \textbf{Order-2}      & \textbf{Order-3}      & \textbf{avg}      & \textbf{Order-4}        & \textbf{Order-5}       & \textbf{Order-6}       & \textbf{avg}     \\ 
\specialrule{1.2pt}{2pt}{2pt}
FT$^*$    & 18.9             & 24.9            & 41.7            & 28.5              & 7.4               & 7.4              & 7.5              & 7.4              \\
LoRA$^*$  &  44.6                &   32.7              &     53.7            &     43.7              &      2.3             &    0.6              &    1.9              &     1.6             \\
LFPT5$^*$            & 67.6             & 72.6            & {77.9}            & 72.7              & 70.4              & {68.2}             & 69.1             & 69.2             \\    
O-LoRA(OIncLoRA)$^*$  & {77.1}               & {76.2}            & 76.6            & {{76.6}}              & {68.4}              & 68.8             &{71.4}             & {\textbf{69.5}}                        \\ 
MoCL$^*$  & {75.6}               & {75.4}            & 76.7            & {75.9}              & {-}              & -             & {-}             & {-}             \\ 
{MoELoRA}  & 52.8               & 49.6           & 59.8           & 54.1             & 36.3              &  31.4            &15.1           & 27.6                        \\ 
\specialrule{1.2pt}{2pt}{2pt}
ProgPrompt$^*$       & 75.2             & 75              & 75.1            & 75.1              & 78.0              & 77.7             & 77.9             & 77.9             \\
PerTaskFT$^*$      & 70.0               & 70.0              & 70.0              & 70.0                & 78.1              & 78.1             & 78.1             & 78.1             \\
MTL$^*$   & 80.0               & 80.0              & 80.0              & 80.0                & 76.5              & 76.5             & 76.5             & 76.5             \\ 
\specialrule{1.2pt}{2pt}{2pt}
FT & 74.4 & 75.0& 77.5& 75.7 & 70.6& 69.7 & 65.6 & 68.3\\
MIGU + FT& 78.3 & \textbf{79.8} & \textbf{78.3}& 78.8 & \textbf{77.1} & 73.6 & 70.7 & 73.8\\
FTReplay & 77.4 & 77.1& 77.9& 77.4  & {-}              & -             & {-}             & {-}             \\ 
MIGU + FTReplay & \textbf{80.8} & 77.0& 78.0& \textbf{78.6}  & {-}              & -             & {-}             & {-}             \\ 

LORA & 60.7 & 70.0 & 73.1 & 67.9 & 53.7 & 44.4 & 39.8& 46.0\\
MIGU + LORA & 74.8 & 71.6 & 73.5 & 73.3 &66.9 & 64.8 & 51.8 & 61.2\\
IncLoRA &         67.0       &  66.7       &     72.6        &     68.8       &     65.5       &    64.9   &    63.9          &    64.7\\
MIGU + IncLoRA  &         77.2      &  76.7       &     75.4        &     76.4       &     71.3       &    67.7   &    67.3          &    68.7\\
OIncLoRA & 77.1 & 76.2 & 76.6 & 76.6 & 68.4& 68.8 & 71.4& 69.5\\
MIGU + OIncLoRA & 77.1& 77.0& 75.6& 76.6& 67.3& 68.5 & 74.2 & 70.0 \\
LORAReplay & 77.1 & 73.4 & 73.2 & 74.5 &74.5&75.4& 75.7&  75.2\\
MIGU + LORAReplay & 77.8&75.1 &75.9&76.2   & 75.4 & \textbf{76.8} & \textbf{77.2} & \textbf{76.5} \\
\specialrule{2pt}{2pt}{\belowrulesep}
\end{tabular}
}
\caption{Summary of the results on two standard CL benchmarks with T5-large model. Averaged accuracy after training on the last task is reported. All results in the last block are averaged over 3 runs. (We reuse the table template and experiment results from O-LoRA~\cite{wang2023orthogonal} to construct the results of the top two blocks, methods denoted with $^*$). \textbf{It is noticeable some baselines in some previous literature show significant lower performance than ours(SeqFT in block 1 v.s. FT in block 3), we assume this may due to different hyperparameter choice for baseline methods. To this end, we conducted a grid search that ranges in [1e-3, 3e-4, 1e-4, 3e-5] for FT baseline. We recorded our experiment results and provide in Table~\ref{tab:appendix_abl_lr}.
 }}
\label{detained_main_results}
\end{table*}

\subsection{Training orders}
The training orders in 3 benchmarks on T5-large and RoBERTa models are shown in table~\ref{table:Training_order}.

\begin{table}
    \centering
    \resizebox{1.0\linewidth}{!}{
    \begin{tabular}{lcccc}
    \toprule
     learning rate & \textbf{Order1} & \textbf{Order2} & \textbf{Order3} & \textbf{Avg} \\
    \midrule
    (Seq)FT in O-LORA & 18.9 &	24.9	& 41.7 &	28.5 \\
    \midrule
    1e-3	& 23.0 & 	21.7 & 	21.8 &	22.2\\
3e-4 &	53.1&	50.8&	31.9&	45.3\\
1e-4&	74.4&	75.0&	77.5&	75.7\\
3e-5&	75.0&	76.1&	77.2&	76.1\\
    \bottomrule
    \end{tabular}}
    \caption{The grid search for baseline FT. It is clearly that learning rates play an important role for CL.}
    \label{tab:appendix_abl_lr}
\end{table}

\section{Baselines for all settings}
\subsection{Baselines on Standard CL benchmark and Long sequence benchmark}\label{baseline_t5_large}

We reuse some baseline descriptions from O-LoRA~\cite{wang2023orthogonal}.

\begin{itemize}[leftmargin=*, align=left]
    \item \textbf{FT} \cite{de2019episodic}: train all model parameters on a sequence of tasks (without adding any regularization or replaying samples from the previous tasks).
    \item \textbf{LoRA}: fixed-size LoRA parameters are trained on a sequence of tasks (without adding any regularization or replaying samples from the previous tasks).
    \item \textbf{IncLoRA}: incremental learning of new LoRA parameters on a sequential series of tasks (without adding any regularization or replaying samples from the previous tasks).
    \item \textbf{Replay}: finetune the whole model with a memory buffer, and replay samples from old tasks when learning new tasks to avoid forgetting.
    \item \textbf{LFPT5} \cite{qin2021lfpt5}: continuously train a soft prompt that simultaneously learns to solve the tasks and generate training samples, which are subsequently used in experience replay.
    \item \textbf{OIncLoRA} \cite{wang2023orthogonal}: learns tasks in different LoRA subspaces that are kept orthogonal to each other and sums all LoRA weights up at testing time.
    \item \textbf{MoCL} \cite{wang2024rehearsalfree}: MoCL continually adds new modules to language models and composes them with existing modules. 
    \item \textbf{SAPT} \cite{zhao2024sapt}: In the SAPT method, a Shared Attentive Learning and Selection Module (SALS) is employed to guide training samples through optimal PET blocks for task-specific learning, using a unique instance-level attention mechanism. This process ensures efficient continual learning for large language models.
    \item \textbf{MoELORA} \cite{luo2024moelora}: MoELoRA considers LoRA as a Mixture of Experts, leveraging the modeling capabilities of multiple experts for complex data domains, as well as utilizing LoRA’s parameter-efficient characteristics.
\end{itemize}

\begin{itemize}[leftmargin=*, align=left]
    \item  \textbf{NCL} (Naive CL) continually DAP-trains the RoBERTa; 
    \item \textbf{NCL-Adapter} continually DAP-trains a set of adapters~\citep{Houlsby2019Parameter}
    \item \textbf{DER++}~\citep{buzzega2020dark} is a replay method based on knowledge distillation. 16.4K tokens are saved for each domain in the replay memory.
    \item \textbf{DEMIX}~\citep{gururangan2021demix} adds a new adapter for each new domain and initializes it with a previous adapter nearest to the new domain; 
    \item \textbf{HAT-Adapter}~\citep{Serra2018overcoming}: HAT is an effective {parameter-isolation} method. HAT is applied to Transformer layers (i.e., self-attention, intermediate and output layers). 
    \item \textbf{ HAT-Adapter}~\citep{ke2021adapting}: HAT-Adapter uses HAT within adapters.
    \item \textbf{DAS}~\cite{ke2023continual} DAS proposes a soft-masking method to overcome CF and to encourage KT, and a constrative learning-based method for knowledge integration. 
\end{itemize}

\section{Experimental Results}
\subsection{Experiment on T5}\label{Experiment_on_T5}
\begin{figure}[h]
    \centering
    \includegraphics[width=0.9\linewidth]{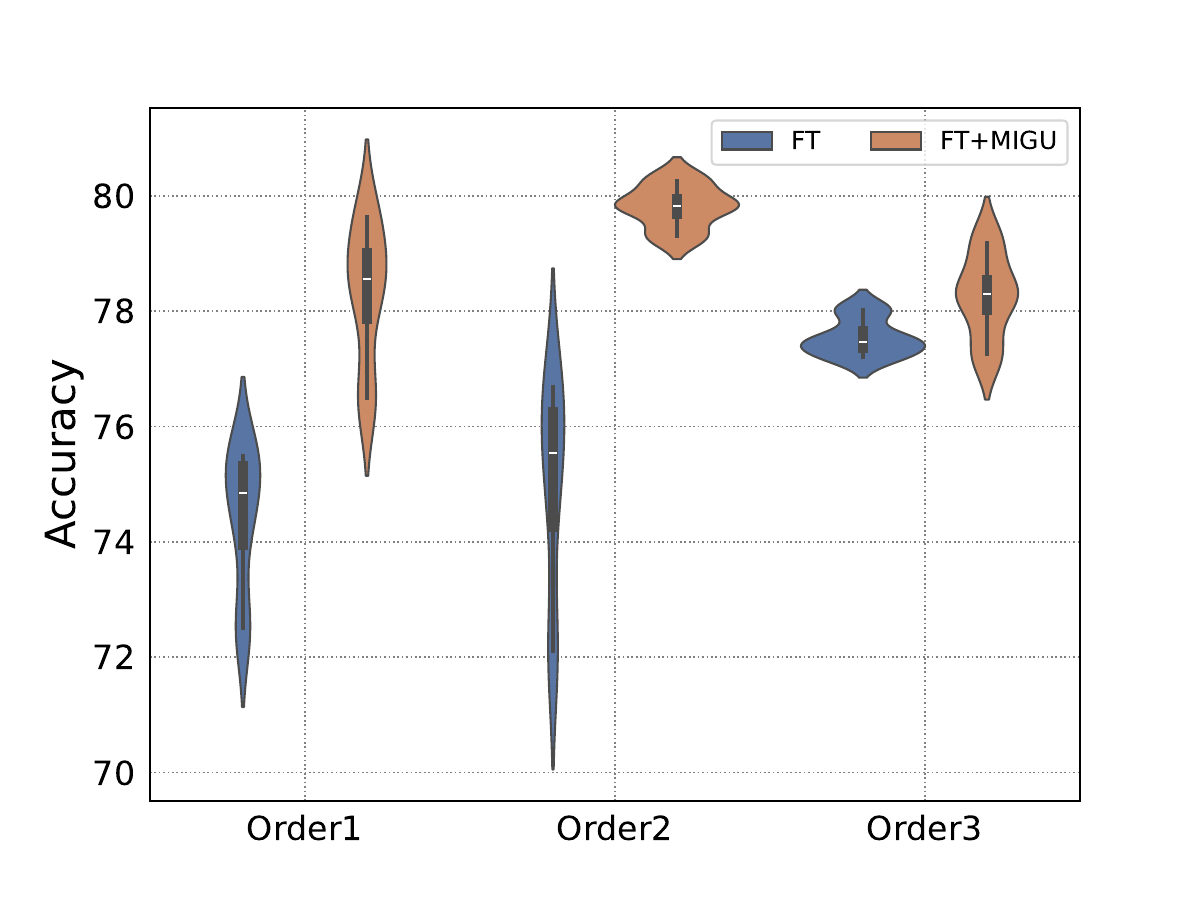}
    \caption{Performance comparison on the standard cl benchmark under full finetuning setting, with and without the implementation of our method..}
    \label{fig:order123}
\end{figure}

\begin{figure}[h]
    \centering
    \includegraphics[width=0.9\linewidth]{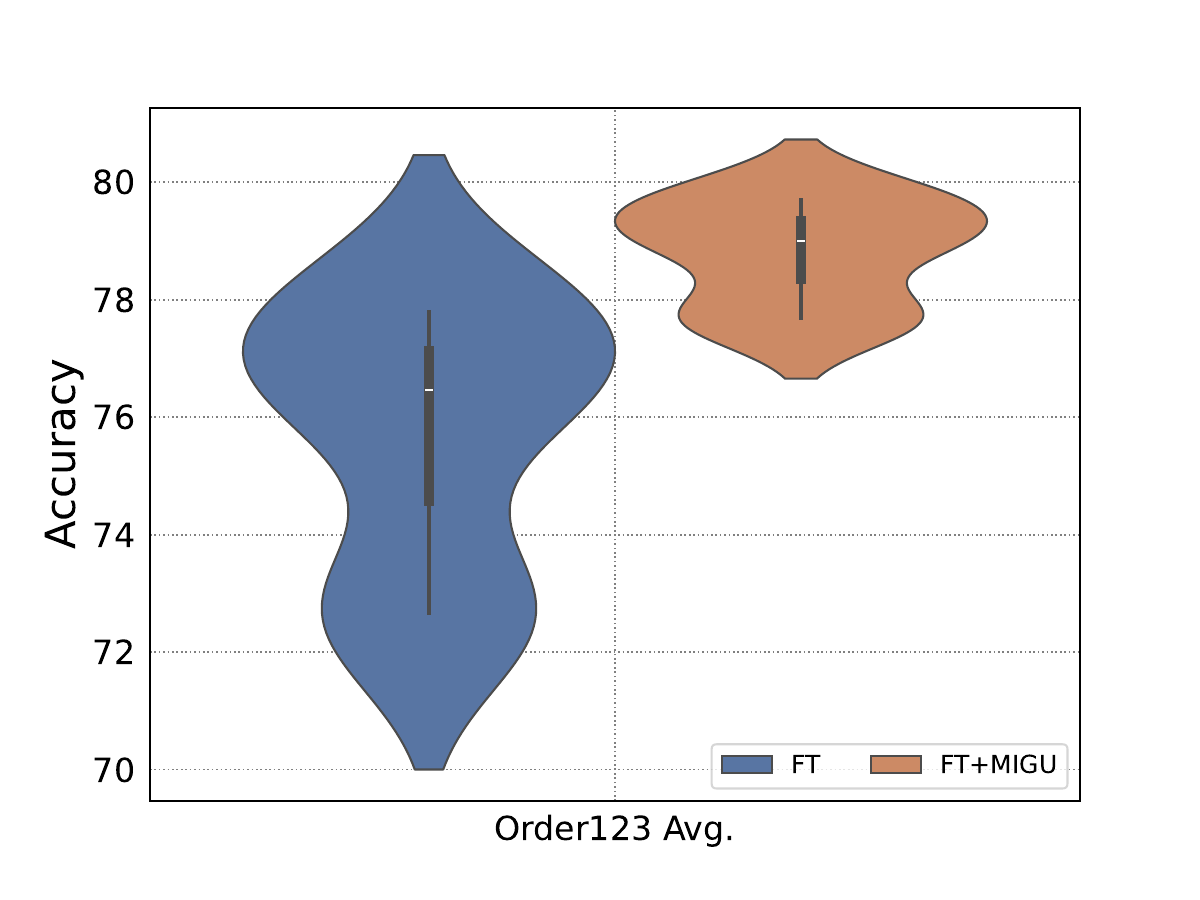}
    \caption{Average performance comparison on the standard cl benchmark under full finetuning setting, with and without the implementation of our method..}
    \label{fig:avgorder123}
\end{figure}

\begin{figure}[h]
    \centering
    \includegraphics[width=0.9\linewidth]{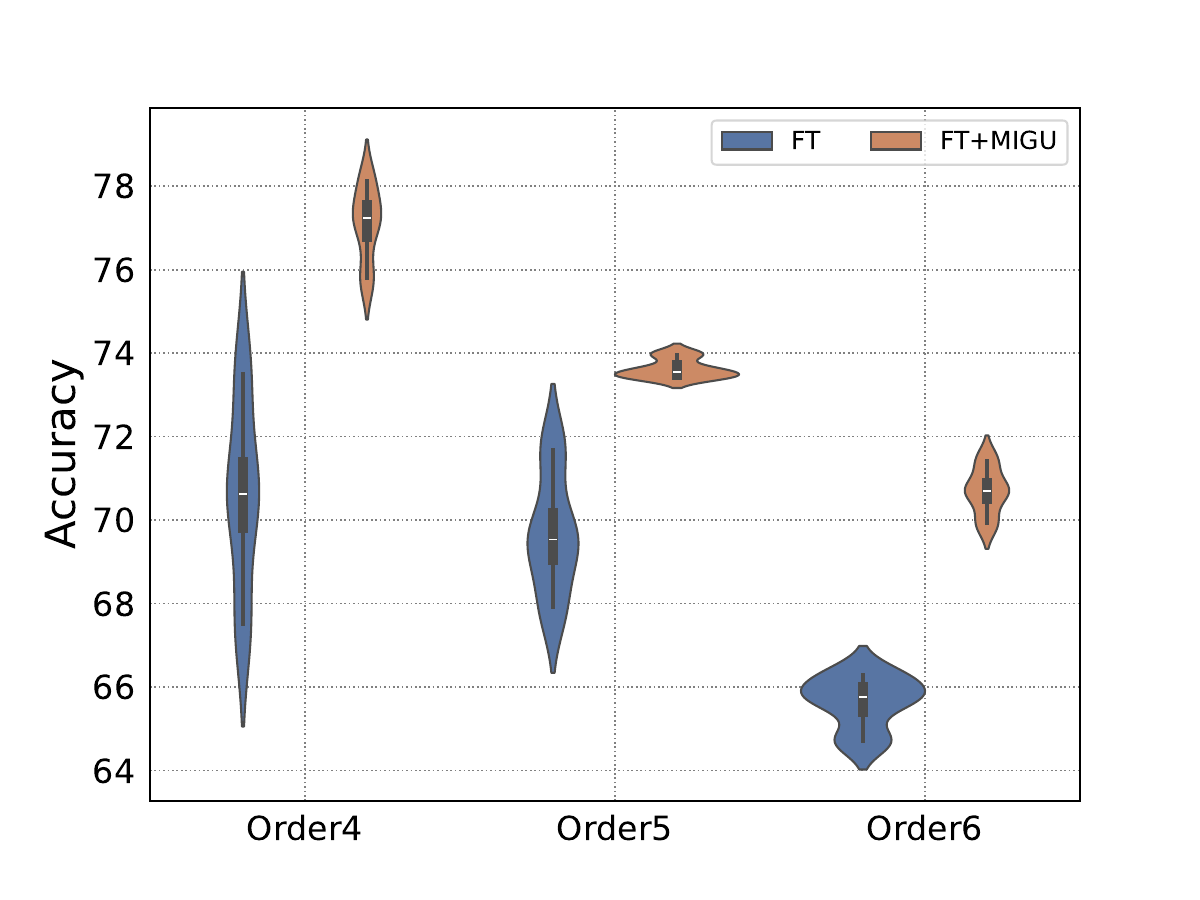}
    \caption{Performance comparison on the standard cl benchmark under full finetuning setting, with and without the implementation of our method..}
    \label{fig:order456}
\end{figure}

\begin{figure}[h]
    \centering
    \includegraphics[width=0.9\linewidth]{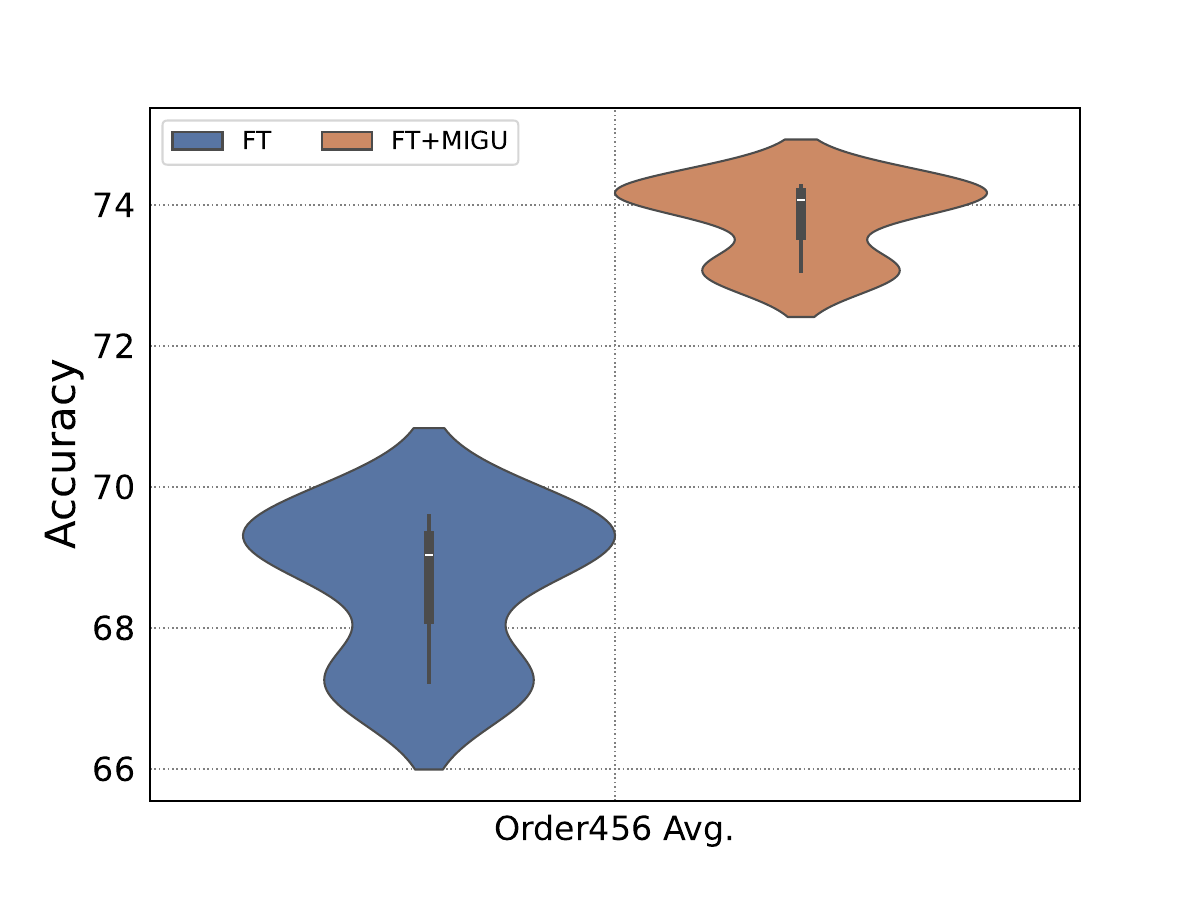}
    \caption{Average performance comparison on the standard cl benchmark under full finetuning setting, with and without the implementation of our method..}
    \label{fig:avgorder456}
\end{figure}

We report more detailed results on the Standard CL benchmark and Long sequence benchmark in table~\ref{detained_main_results}, including each order results and their corresponding average results. 
To more intuitively display our results compared to the baseline, we plotted violin graphs showing the performance with and without our method under the condition of full finetuning as Figure~\ref{fig:order123}~\ref{fig:avgorder123}~\ref{fig:order456}~\ref{fig:avgorder456}.

\subsection{Experiment on RoBERTa}\label{das_detailed}
\begin{figure}[ht]
    \centering
    \includegraphics[width=\linewidth]{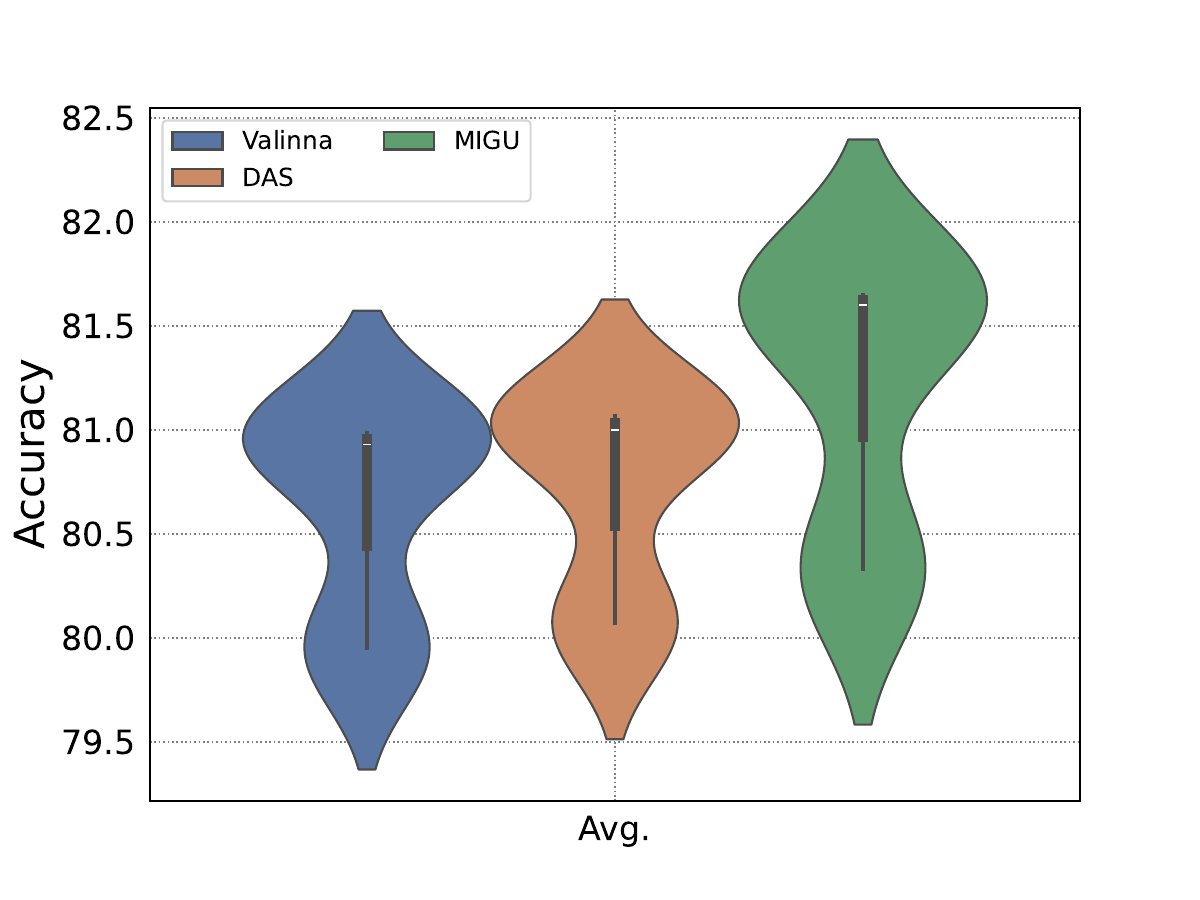}
    \caption{Performance comparison on the DAS benchmark under continual domain pre-training setting, with and without the implementation of our method.}
    \label{fig:das_violin}
\end{figure}

\paragraph{Detailed experiment results}
The violin graphs results is shown as Figure~\ref{fig:das_violin}.

\begin{figure}[th]
    \centering
    \includegraphics[width=1\linewidth]{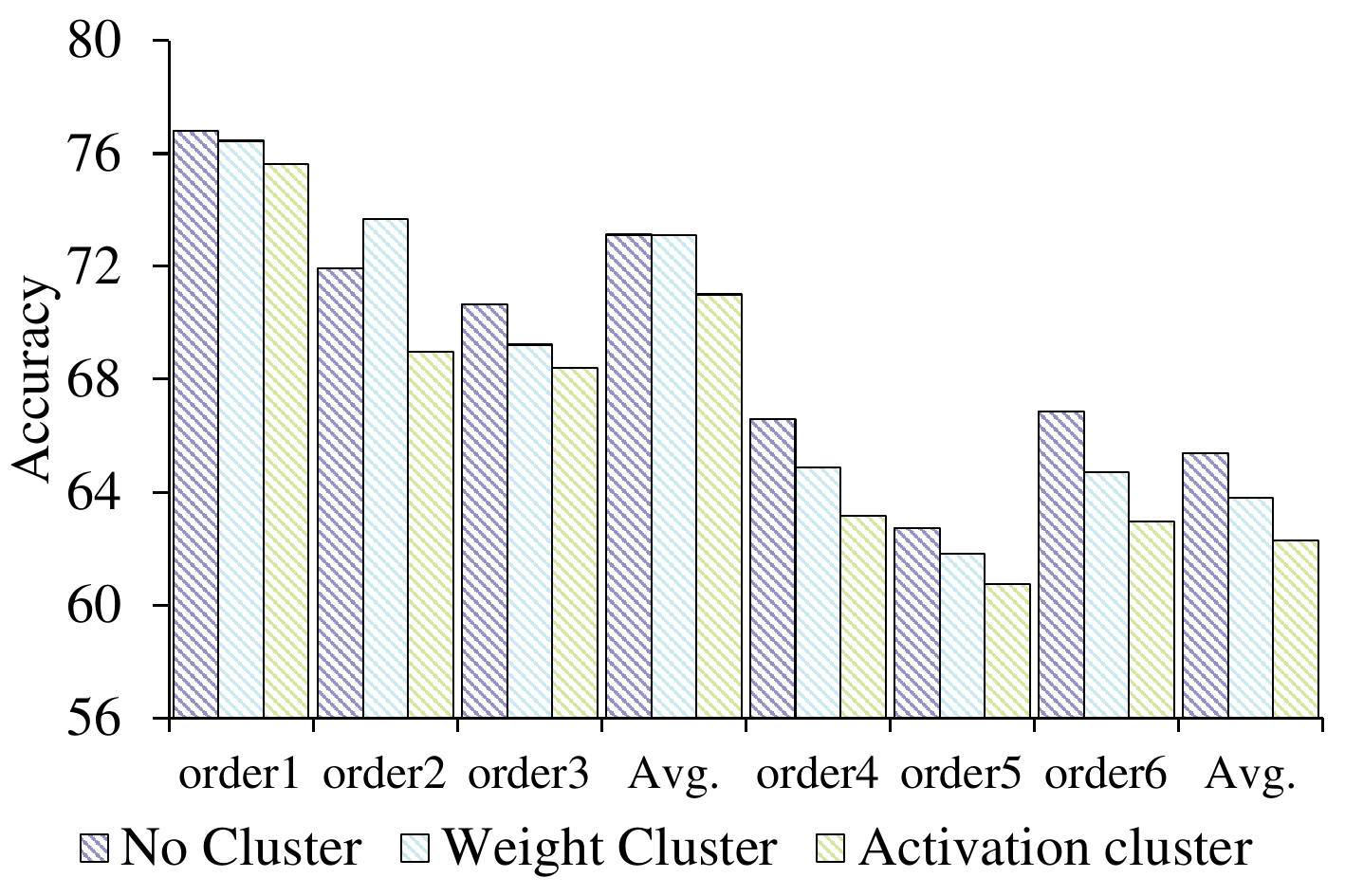}
    \caption{Ablation on LoRA MIGU+cluster design}
    \label{fig:ablation_design}
\end{figure}

\begin{figure*}[th]
    \centering
        \begin{minipage}{1\linewidth}
       \subfigure[ARC-Challenge]{
        \includegraphics[width=0.32\linewidth]{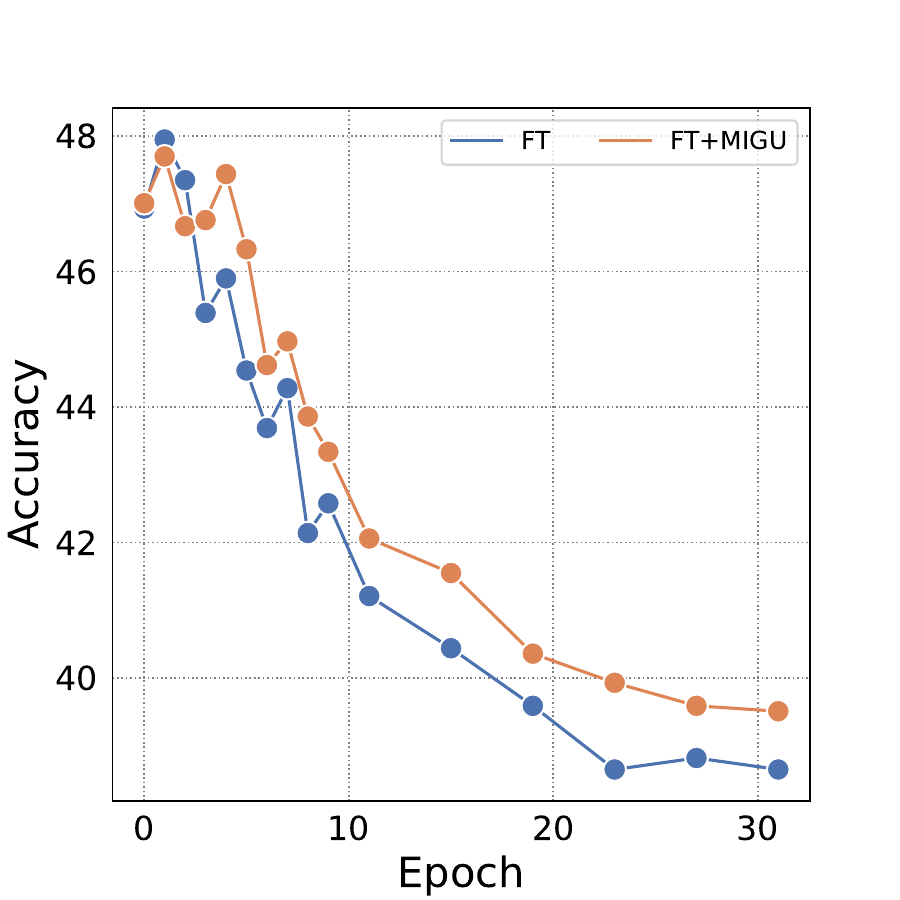}\label{fig:Llama_code_arcc}
        }\noindent
         \subfigure[HellaSwag]{
        \includegraphics[width=0.32\linewidth]{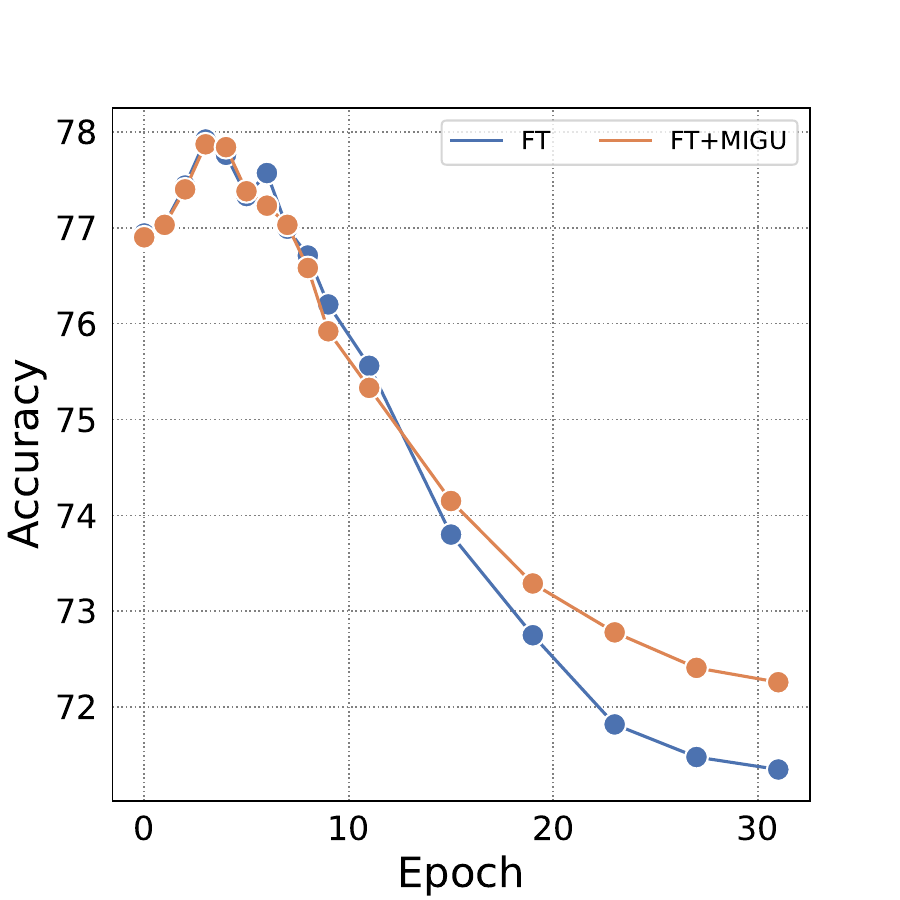}\label{fig:Llama_code_hellaswag}
        }\noindent
       \subfigure[Winogrande]{
        \includegraphics[width=0.32\linewidth]{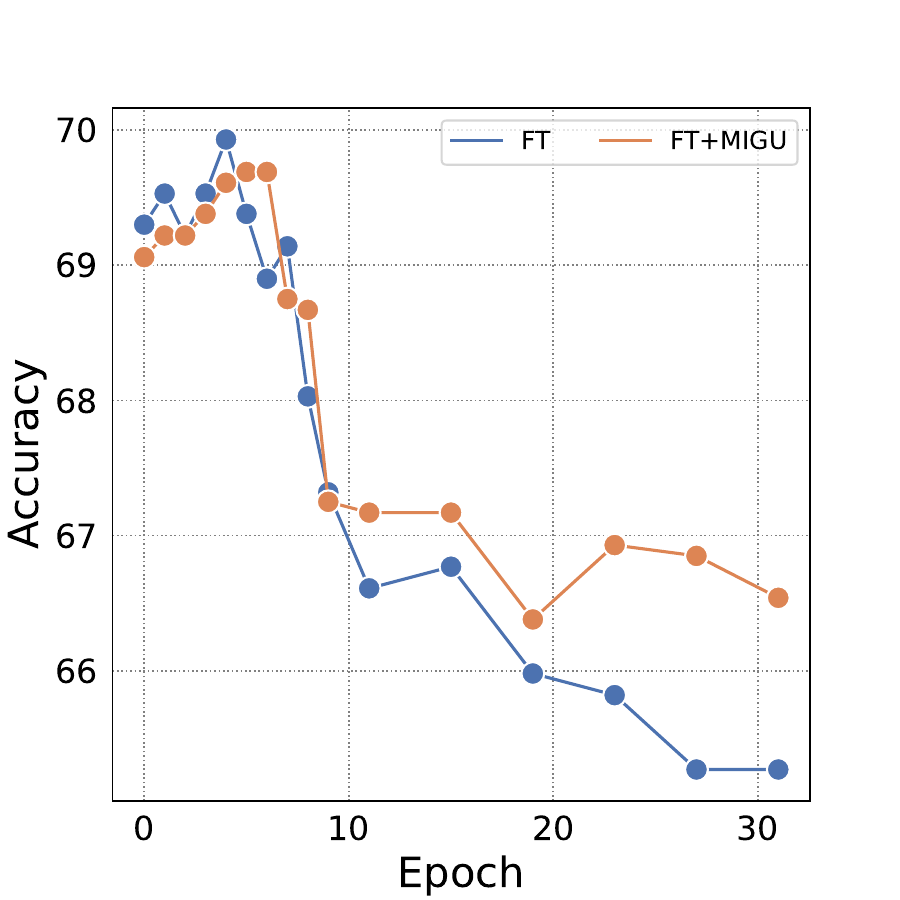}\label{fig:Llama_code_winogrande}
        }
    \end{minipage}
    \caption{Accuracy on ARC-Challenge~\cite{clark2018think}, HellaSwag~\cite{zellers2019hellaswag} and Winogrande~\cite{sakaguchi2019winogrande}, evaluating on Llama-2-7B by MIGU with LoRA and valinna LoRA instruct tuning pre-trained on Magicoder-Evol-Instruct-110k~\cite{wei2024magicoder}.}
    \label{fig:visualization_chose1}
\end{figure*}

\begin{figure}[h]
    \centering
    \includegraphics[width=1\linewidth]{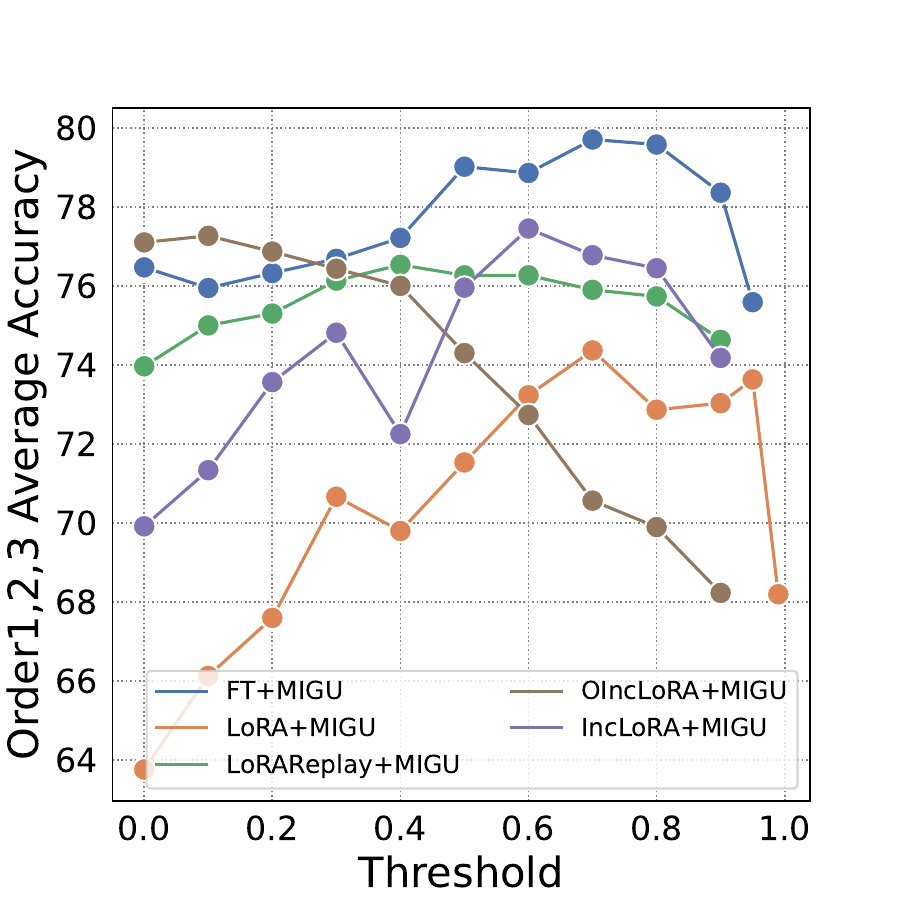}
    \caption{Ablation study on the gradient mask threshold. The curves illustrate that the optimal value is concentrated around 0.7 for FT+MIGU, LoRA+MIGU, and IncLoRA+MIGU, 0.4 and 0.1 for the LoRAReplay+MIGU and OIncLoRA+MIGU settings respectively.}
    \label{fig:threshold}
\end{figure}

\paragraph{Efficiency}
We also conduct an efficiency ablation by comparing FT, FT+MIGU and DAS because continual-pre-training is a relatively computational-intensive setting. 
DAS is a typical parameter-based regularization methods. 
We record the wall time required for the first three dataset given the same GPU configuration: A100 $\times$ 2. 
As shown in the Table~\ref{tab:efficiency}, FT+MIGU only occur an approximately 10\% overhead in wall time, due to the extra masking step in the backward propagation phase while DAS achieves a magnitude larger overhead.

\begin{table}
    \centering
    \setlength{\tabcolsep}{4pt}
    \resizebox{1.0\linewidth}{!}{
    \begin{tabular}{lccc}
    \toprule
     & \textbf{Restaurant} & \textbf{ACL} & \textbf{AI} \\
    \midrule
    FT    &  25.3\small{(+0.0\%)} &  26.7\small{(+0.0\%)} & 15.5\small{(+0.0\%)} \\
    \midrule
     FT + MIGU & 27.3\small{(+7.9\%)}  & 29.2\small{(+9.4\%)} & 17.0\small{(+12.9\%)} \\
     \midrule
    DAS  &  78.0\small{(+208\%)}  & 66.5\small{(+154\%)} &45.0\small{(+190\%)} \\
    \bottomrule
    \end{tabular}}
    \caption{The wall time($\min$) on three domain pre-training dataset.}
    \label{tab:efficiency}
\end{table}

\begin{figure*}[ht]
    \centering
    \includegraphics[width=0.95\linewidth]{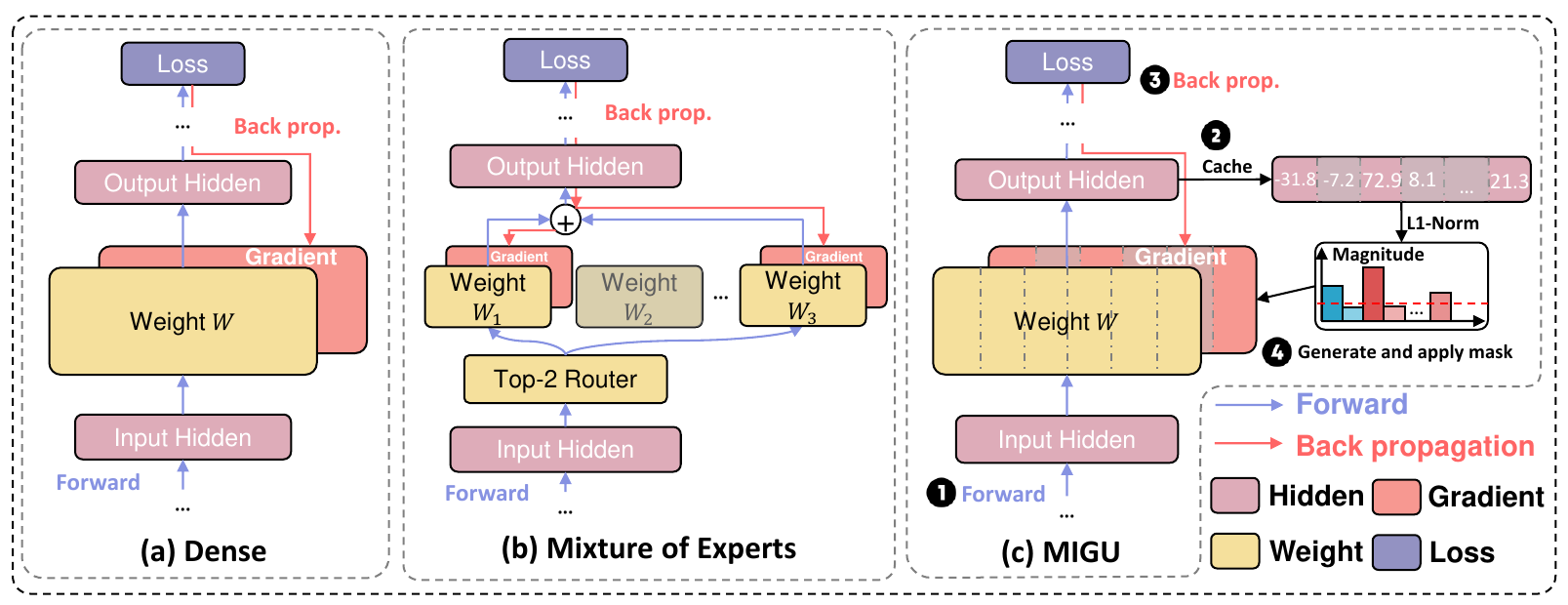}
    \caption{Differences between (a)Dense, (b)MoE and (c)MIGU}
    \label{fig:moe}
\end{figure*}

\subsection{Experiment on Llama2}
The detailed violin graphs results about ARC-Challenge~\cite{clark2018think}, HellaSwag~\cite{zellers2019hellaswag} and Winogrande~\cite{sakaguchi2019winogrande} are seperately shown in Figure~\ref{fig:Llama_code_arcc},~\ref{fig:Llama_code_hellaswag},~\ref{fig:Llama_code_winogrande}.

\subsection{Visualization}\label{visualization}
\begin{figure*}[th]
    \centering
    \begin{minipage}{1\linewidth}
        \subfigure[FT]{
        \includegraphics[width=0.22\linewidth]{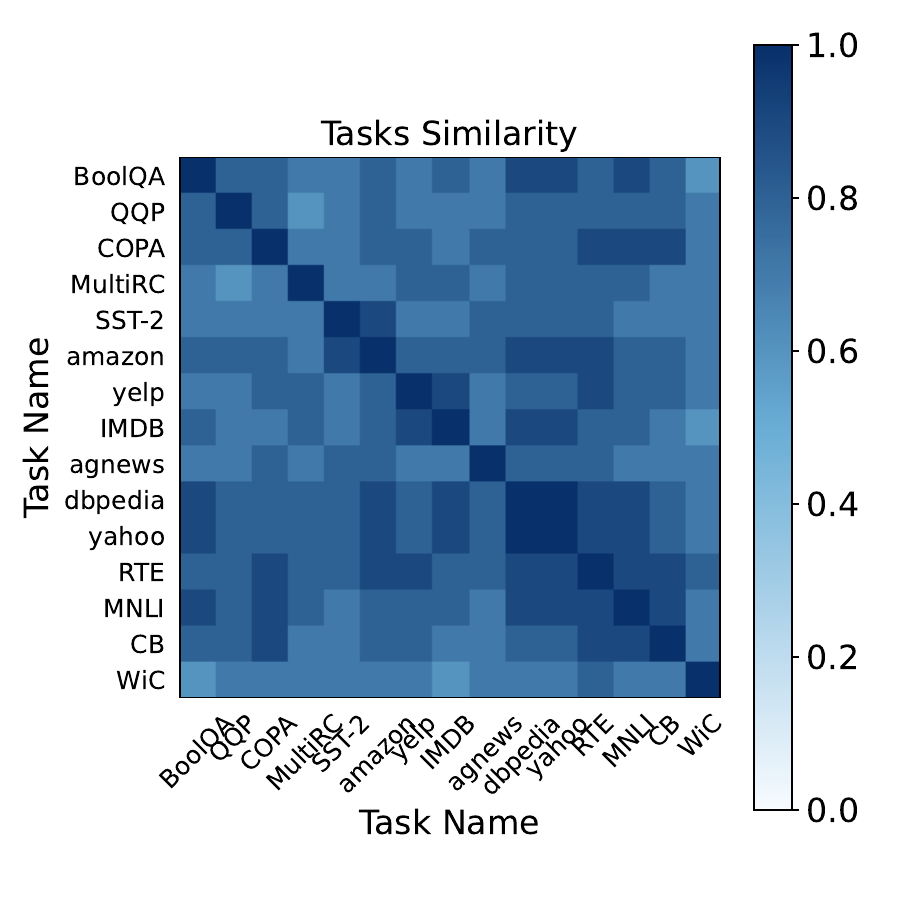}
        }\noindent
         \subfigure[FT + MIGU]{
        \includegraphics[width=0.22\linewidth]{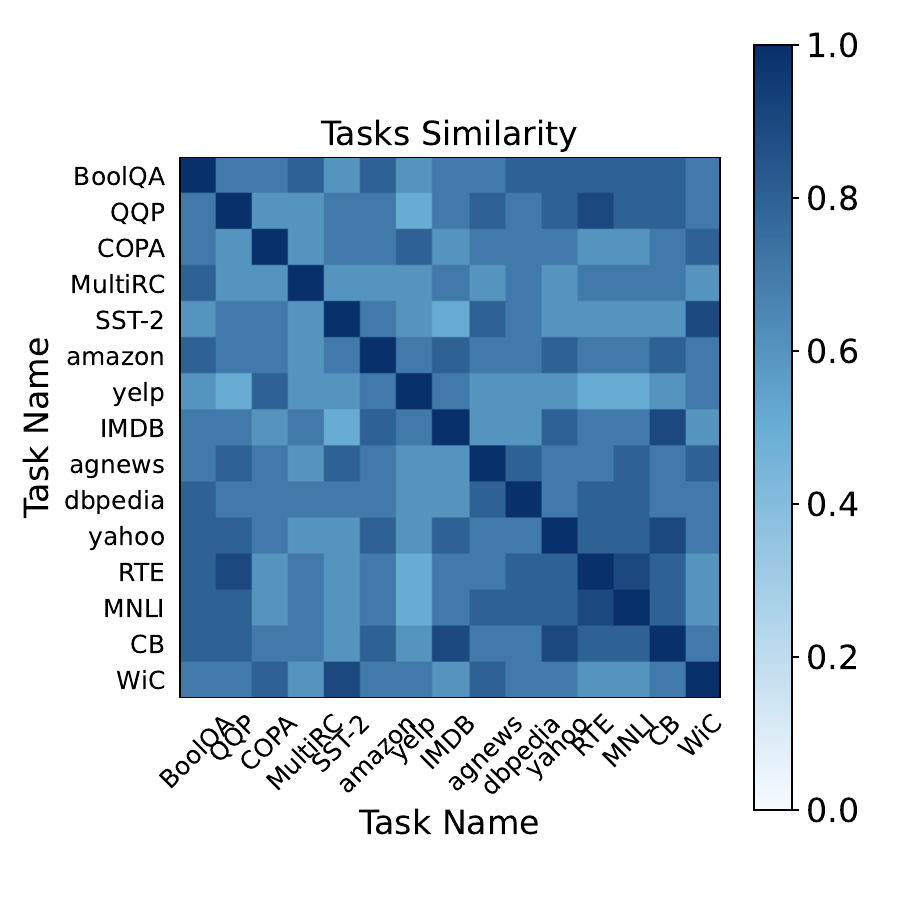}
        }
       \subfigure[FT]{
        \includegraphics[width=0.22\linewidth]{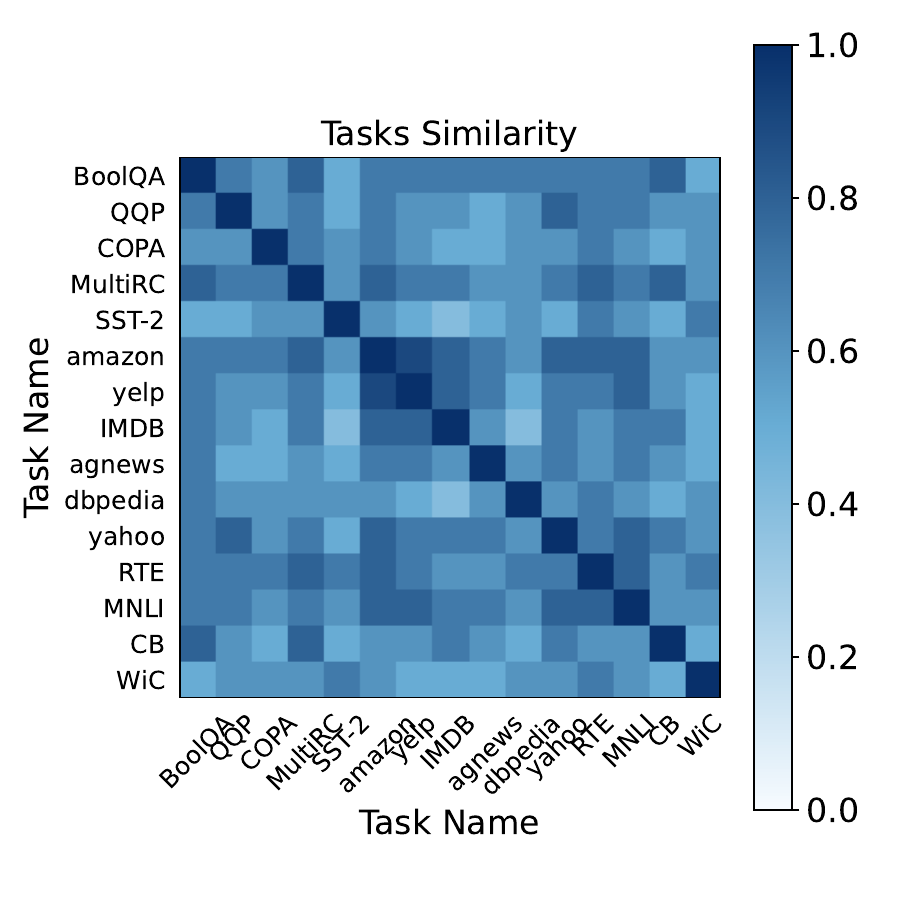}
        }\noindent
         \subfigure[FT + MIGU]{
        \includegraphics[width=0.22\linewidth]{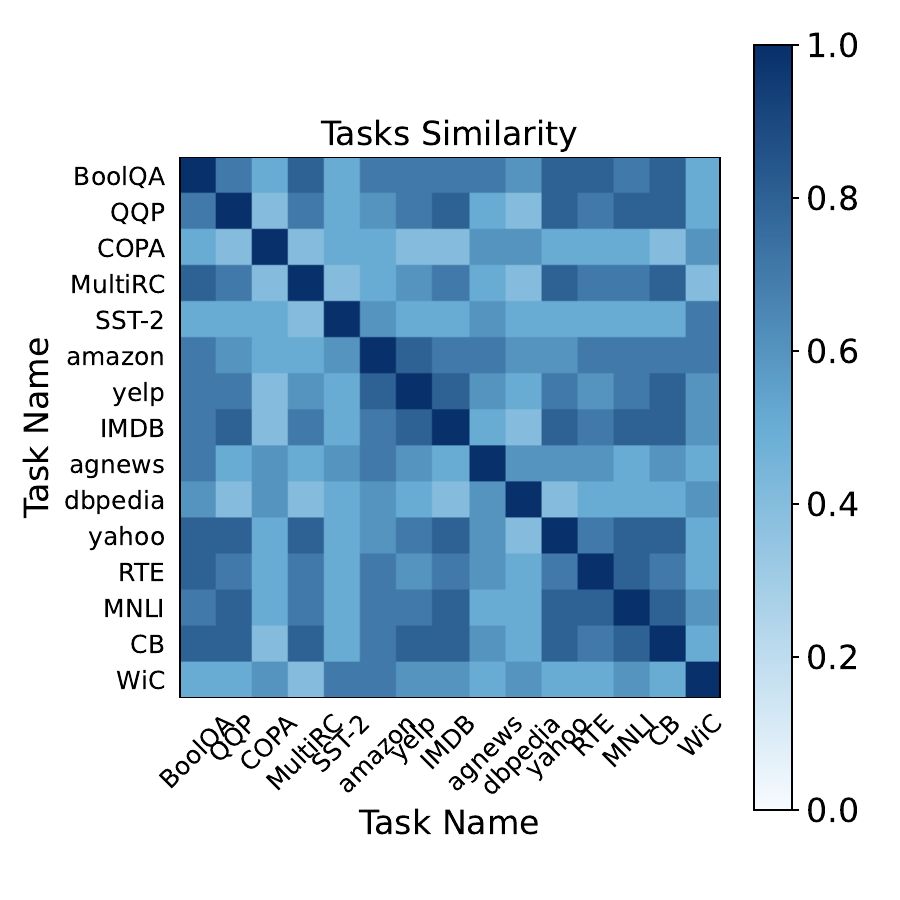}
        }
    \end{minipage}
    \begin{minipage}{1\linewidth}
       \subfigure[FT]{
        \includegraphics[width=0.22\linewidth]{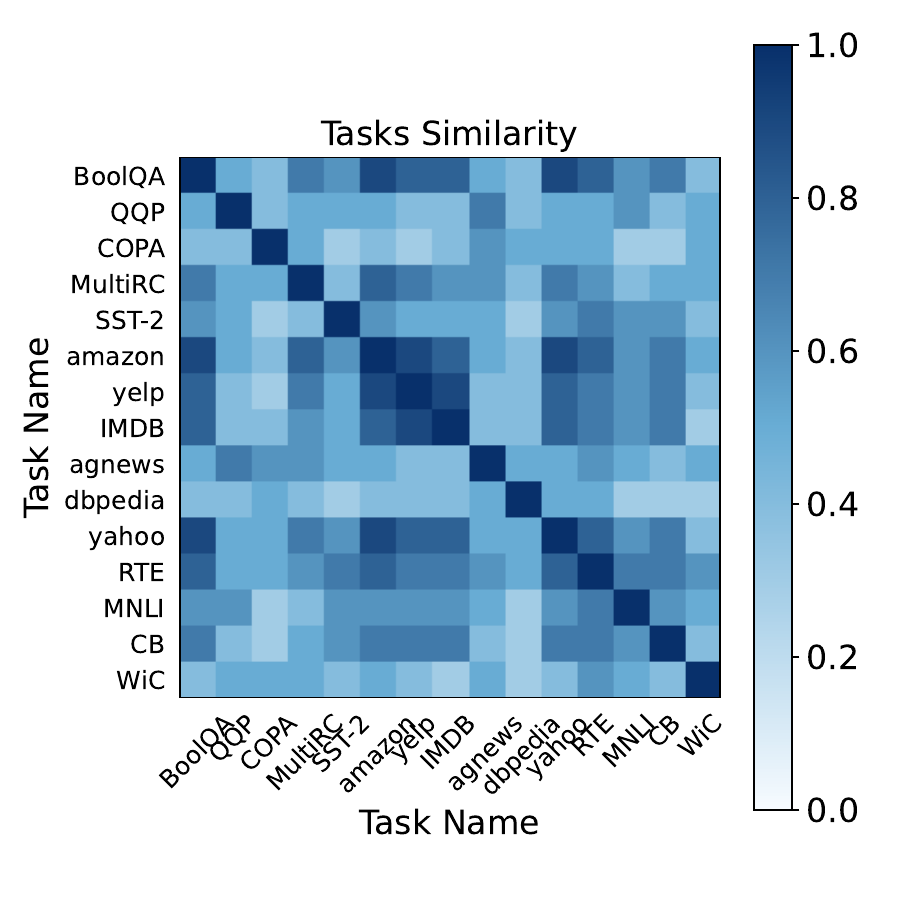}
        }\noindent
         \subfigure[FT + MIGU]{
        \includegraphics[width=0.22\linewidth]{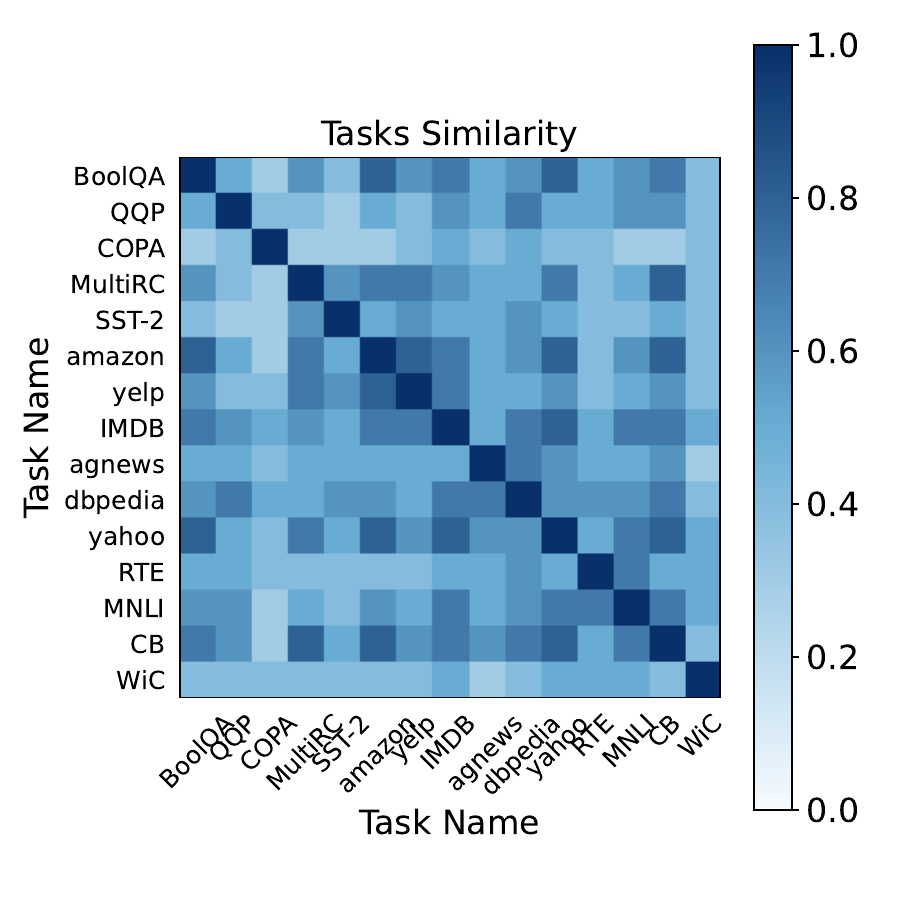}
        }
       \subfigure[FT]{
        \includegraphics[width=0.22\linewidth]{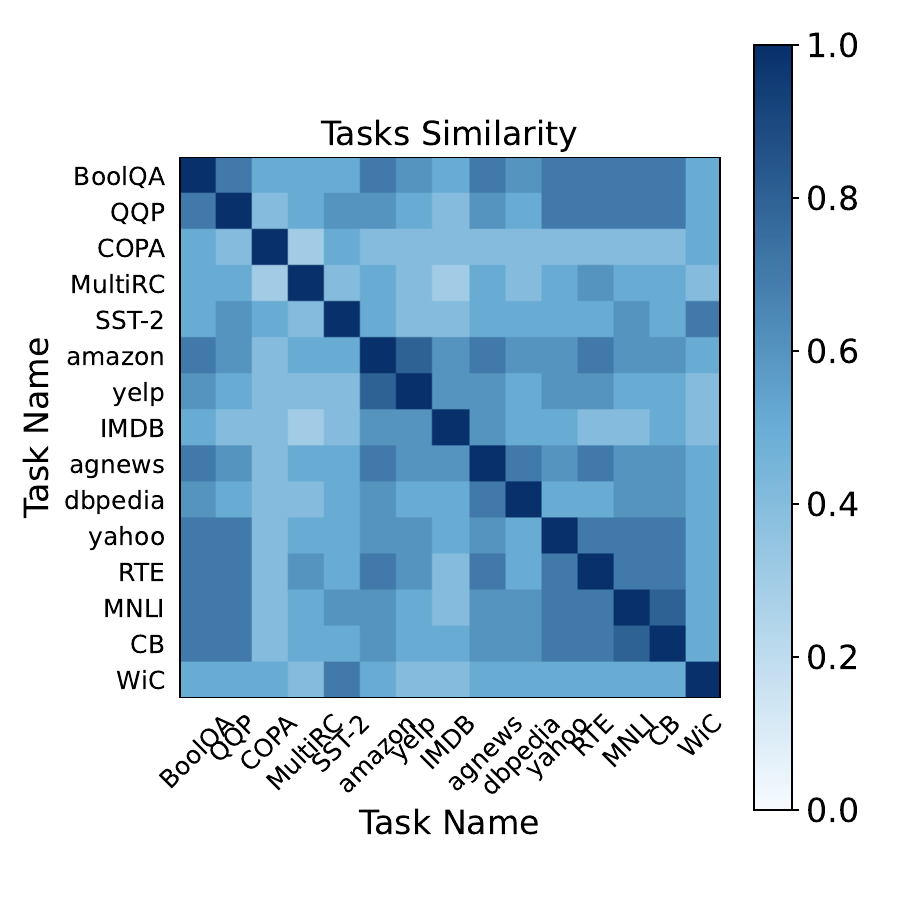}
        }\noindent
         \subfigure[FT + MIGU]{
        \includegraphics[width=0.22\linewidth]{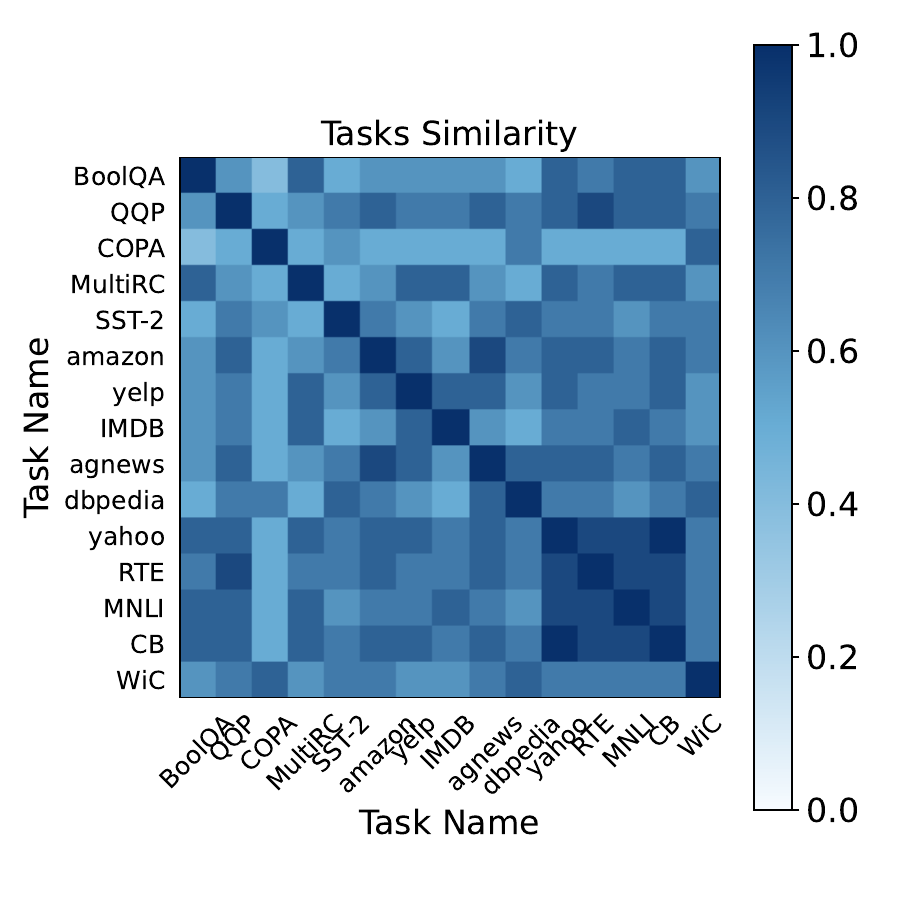}
        }
    \end{minipage}
    \caption{The product magnitude distribution similarity of different tasks in the MHA of the last transformer block: (a,b) query linear layer; (c,d) key linear layer; (e,f) value linear layer; (g,h) output linear layer.}
    \label{fig:visualization_chose2}
\end{figure*}

To investigate how our method enhances model performance, we visualized the variation in product magnitudes between an FT model and an FT model augmented with our MIGU technique in Figures~\ref{fig:visualization_chose1},\ref{fig:visualization_chose2}. We employed heatmaps to depict the similarity in product magnitude distributions across different tasks. Our findings reveal that task similarity in the FT model with MIGU implementation is markedly reduced. This suggests that the models trained with our method exhibit more distinctive weight activations for different tasks, thereby mitigating their conflict. This distinction in activation patterns indicates our method's ability to foster more task-specific representations within the model, contributing to its improved performance across varied learning scenarios.

\paragraph{Magnitude Distribution.}

\begin{figure*}[th]
    \centering
    \includegraphics[width=\textwidth]{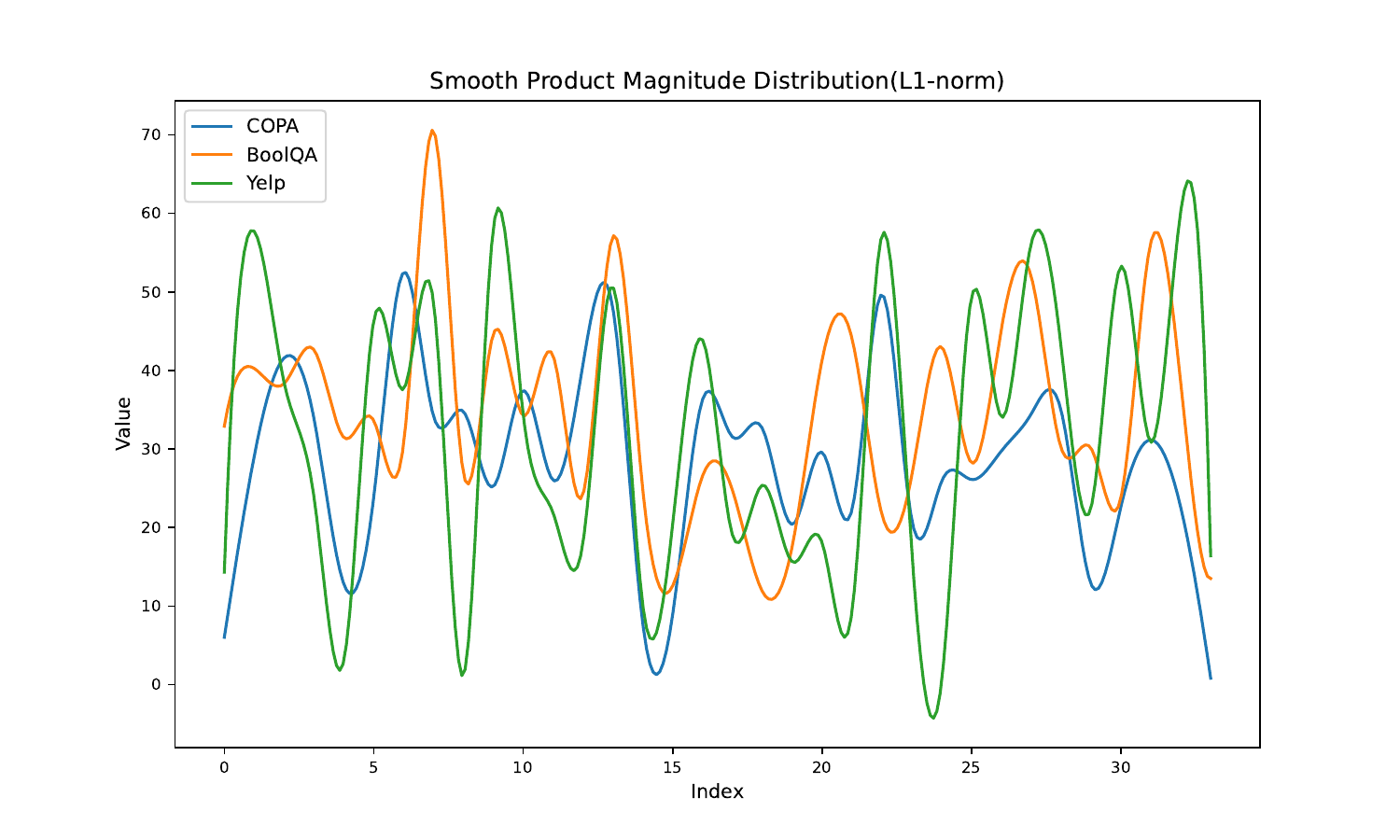}
    \caption{The Magnitudes distribution of COPA sample, BoolQA sample, and Yelp sample on the first linear layer of 23-th FFN layer of T5-large model.}
    \label{fig:smooth_vector_plot}
\end{figure*}

We plot the L1-normalized magnitude distribution of COPA sample, BoolQA sample, and Yelp sample on the first linear layer of 23-th FFN layer of T5-large model in Figure~\ref{fig:smooth_vector_plot}.

\section{Ablation on MIGU+Cluster}

Informed by the works on Mixture of Experts (MoE)~\cite{jiang2024mixtral}, Emergent MoE (EMoE)~\cite{qiu2023emergent}, and MoE\textit{fication}~\cite{zhang2022moefication}, we investigate explicit clustering of weight vectors in LMs to construct expert groups. Technically, we treat the linear layer's weight matrix $\mathbf{W}$ as a set of $d_{out}$ vectors, each of dimension $d_{in}$. These vectors are then partitioned into $N$ clusters, analogous to MoE experts.

\subsection{Implementation}
As detailed in \S~\ref{subsec:MIGU}, our method encompasses four core processes in cluster-based implementation.
 During the data forward phase, the product magnitudes of the weight vectors are computed and tracked. 
 Subsequently, in the second phase, MIGU caches these magnitudes and employs an L1-norm normalization to derive a gradient mask. This mask is pivotal for modulating the gradients in the subsequent phases. 
 The third phase involves the standard backpropagation to calculate the gradients of the parameters. 
 Finally, in the fourth phase, the earlier computed gradient mask is applied to the obtained gradients, ensuring a modulated update of the parameters. This modulation is consistent within each cluster, thereby maintaining the integrity of the expert groupings and enhancing the model's learning efficacy.
We also plot a Figure~\ref{fig:moe} to illustrate the differences between Dense, MoE and ours in forward and backward phase.

We explored two distinct clustering strategies:
\begin{itemize}
    \item Weight Cluster Combination: The weight vectors are clustered into $N$ groups based on their proximity in the weight space.
    \item Co-magnitude Guided Combination: Using a subset of the dataset, we group weight vectors into clusters based on the similarity of their product magnitudes.
\end{itemize}
 \subsection{Result \& Analysis}
The outcomes of two distinct clustering approaches, alongside our implementation within LoRA, are illustrated in Figure~\ref{fig:ablation_design}. It is evident that, except for the second order, the ``Weight Cluster'' method surpasses the 'No Cluster' approach, which does not employ explicit clustering. However, the 'No Cluster' method demonstrates superior performance across the remaining orders, highlighting its robustness and effectiveness. Nonetheless, the other two explicit clustering techniques still significantly outperform the baseline vanilla continual learning LoRA, indicating their potential for further exploration.

\begin{figure*}[th]
    \centering
        \begin{minipage}{1\linewidth}
       \subfigure[FT]{
        \includegraphics[width=0.22\linewidth]{figs/task_sim_full/FT/23L-wi.pdf}
        }\noindent
         \subfigure[FT + MIGU]{
        \includegraphics[width=0.22\linewidth]{figs/task_sim_full/FT_ours/23L-wi.pdf}
        }\noindent
       \subfigure[FT]{
        \includegraphics[width=0.22\linewidth]{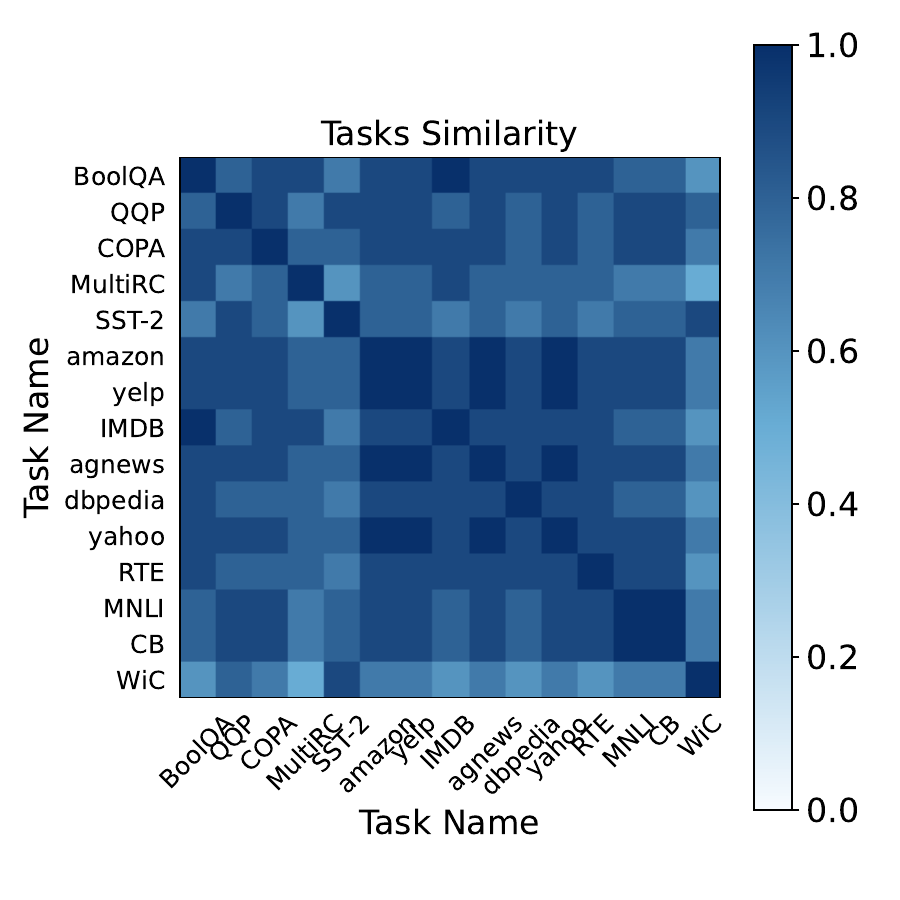}
        }\noindent
         \subfigure[FT + MIGU]{
        \includegraphics[width=0.22\linewidth]{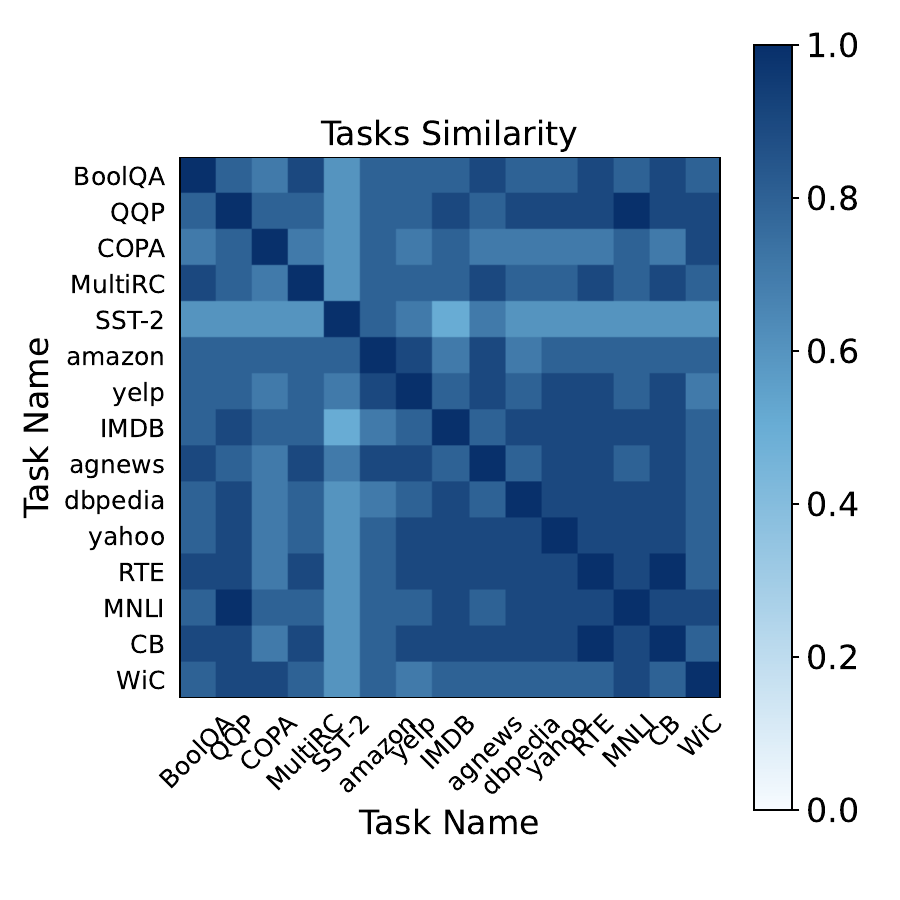}
        }
    \end{minipage}
    \caption{The product magnitude distribution similarity of different tasks in the FFN of the last transformer block: (a,b) 1-st linear layer; (c,d) 2-nd linear layer.}
    \label{fig:visualization_chose1}
\end{figure*}

\begin{table}
    \centering
    \resizebox{1.0\linewidth}{!}{
    \begin{tabular}{lcc}
    \toprule
      & \textbf{Standard Benchmark} & \textbf{Long Benchmark} \\
    \midrule
    FT & 76.5 & 69.6\\
     + gradient-based  & 78.4 &	72.3\\
     + weight-based  & 77.2 &	71.3\\
     + MIGU   & \textbf{78.8} &	\textbf{74.7}\\
    \bottomrule
    \end{tabular}}
    \caption{Finding Important Weights Comparison}
    \label{tab:abl_gradient_importance_method}
\end{table}

\section{More Ablation on Gradient Mask Threshold}\label{more_threshold}

We also used the original standard CL benchmark and plot all five curves of our approach (+MIGU) for gradient mask threshold from 0.0 to 0.9 in Section~\ref{CIT_T5}. 
The optimal threshold value for FT+MIGU, LoRA+MIGU, and IncLoRA+MIGU settings is 0.7 while LoRAReplay+MIGU is 0.4 as shown in Figure~\ref{fig:threshold}\footnote{The ablation on threshold search only reports one run, so it does not align with the results in Section \ref{CIT_T5}.}. OIncLoRA+MIGU is only 0.1, which may due to the parameter updating regularized by the OIncLoRA method itself. The optimal value for IncLoRA+MIGU is 0.6, close to FT+MIGU, LoRA+MIGU, and IncLoRA+MIGU settings. 
Surprisingly, with only 5\% ($T=0.95$) or 1\% ($T=0.99$) parameters updating, LoRA+MIGU still beats LoRA by a wide margin. 
This interesting finding may indicate that only a small proportion of proportional weights with large magnitudes is crucial for successful CL settings, which may be worth future investigation.

\end{document}